\documentclass{article}
\usepackage[margin=1.25in]{geometry}
\usepackage{graphicx} 
\usepackage{amsmath}
\usepackage{amsfonts}
\usepackage{amssymb}
\usepackage{booktabs}
\usepackage{adjustbox}
\usepackage{cellspace}
\usepackage{wrapfig}
\usepackage{amsthm}
\usepackage{mathtools}
\usepackage{csquotes}
\usepackage{array}
\usepackage{url}
\usepackage{natbib}
\usepackage{authblk}
\usepackage{parskip}
\usepackage[dvipsnames]{xcolor}

\newtheorem{theorem}{Theorem}
\newtheorem{proposition}{Proposition}
\newtheorem*{explanation}{Explanation}

\DeclareMathOperator*{\argmax}{arg\,max}
\DeclareMathOperator*{\argmin}{arg\,min}
\DeclareMathOperator\supp{supp}
\DeclareMathOperator\image{Im}

\usepackage{soul}
\usepackage{xcolor} 
\newcommand{\eric}[1]{{\color{red}{\bf\sf [Eric: #1]}}}

\title{Critique of World Model
}

\author{
  Eric Xing\textsuperscript{$\diamond$,†}\thanks{~~Co-first author}\ ,
  Mingkai Deng\textsuperscript{$\diamond$,†}$^*$,
  Jinyu Hou\textsuperscript{$\diamond$,†}
  \\
  \textsuperscript{$\diamond$}Institute of Foundation Models, Mohamed bin Zayed University \\ of Artificial Intelligence \\
  \textsuperscript{†} School of Computer Science, Carnegie Mellon University 
  \\~\\
  \{eric.xing, mingkai.deng, jinyu.hou\}@mbzuai.ac.ae
}

\date{June 1, 2026}

\begin{document}

\maketitle

\begin{abstract}
World Model, the algorithmic simulator  of the real-world environment which biological agents experience and act upon, has been an emerging topic in recent years due to the rising need to develop virtual agents with artificial (general) intelligence. There has been much discussion on what a world model really is, how to build it, how to use it, and how to evaluate it. In this essay, starting from the imagination in the famed Sci-Fi classic Dune, and drawing inspiration from the concept of ``hypothetical thinking'' in psychology literature, we argue the primary goal of a world model to be {\it simulating all actionable possibilities of the real world for purposeful reasoning and acting}. We examine the key design dimensions of world modeling: data, representation, architecture, learning objective, and usage, surveying existing approaches and analyzing their tradeoffs. Building on this examination, we propose a new Generative Latent Prediction (GLP) architecture for a general-purpose world model, based on 
stateful,
hierarchical, multi-level, and mixed continuous/discrete representations, and a generative and self-supervised learning framework, with an outlook of a Physical, Agentic, and Nested (PAN) AGI system enabled by such a model.
\end{abstract}

\section{Introduction}
A Large Language Model (LLM) simulates the \textbf{next word} in human languages, which has led to systems like ChatGPT that allow people to perform a wide range of tasks facilitated through language, such as common conversation, standardized tests, professional writing, and advanced math reasoning, at a level rivaling human intelligence.

What would you do if you could perfectly simulate the \textbf{next world} -- every possible future in the environment that we reside? \textit{Dune}, a classic of science fiction that inspired the likes of George Lucas' \textit{Star Wars} and Miyazaki's \textit{Valley of the Wind}, boldly imagines such a possibility. The series is centered around the \textit{Kwisatz Haderach}, a prophesized human being who inherits their ancestors' memories and simulates the outcomes of all possible plans in order to chart the best path to achieve their goals~\cite{mcnelly1984dune}. Such superhuman ability allows them to command armies to win galactic wars, or to oversee global-scale projects that turn a desert planet into a green paradise. 
Is it possible to build towards computer systems with similar functionalities using a similar approach? 

Unlike a chat-bot, human consists of a hierarchy of abilities that go from immediate, concrete ones (e.g., body control/movement/action, reading/listening, and speaking/drawing) to far-reaching and abstract ones (e.g., planning, collaboration, and strategizing). Furthermore, the same human, while not necessarily perfectly, can perform a broad range of tasks (e.g., do household chores, complete risky expeditions, conduct research investigations, navigate social situations, and manage complex enterprises) all with the same cognitive architecture of the human brain. Can a single artificial intelligence (AI) system perform all these tasks? Each of these problems can be seen as a goal-oriented agent acting in a multimodal environment, requiring purposeful reasoning of massive temporal-spacial, social-physical, and emotional-cognitive complexity and depth, to the point that traditional approaches built on logical inference (e.g., induction, deduction, abduction) are often easily overwhelmed. It emerges that the key to generalized decision-making toward such complexity lies in reasoning by ``seeing the future'' as
seen in \textit{Dune} -- an ability formally known as \textit{Hypothetical Thinking}~\cite{Ball_2020} in the psychology literature, or ``thought experiments" in common practice -- the ability to simulate the next worlds using a mental model of the world. We call such a mental model a \textbf{World Model}. 

Specifically, a World Model (WM) is a generative model that simulates the possibilities in diverse scenarios (e.g., physical world, mental world, social world, and evolutionary world). Operationally, a WM takes previous world state $s$ and action $a$, and predicts or simulates the next world state $s'$ through a transformation function, such as a conditional probability distribution:
\begin{equation}
    s' \sim p(s' | s,a)
\end{equation}

With a WM, machines can perform thought experiments by simulating actions and plans in complex scenarios, including counterfactual ones, and extracting the best ones among them. This is consistent with the hypothesis that humans reason not just linearly with formal rules toward goals (e.g., imagine a self-serving individual immediately offering help to another person upon seeing them cry in the hopes of stopping them from crying) via a deterministic optimization algorithm, but also based on simulations using an internal mental model (e.g., imagine the same individual decides what to do by mentally \textit{simulating} multiple possible outcomes including self-exhaustion, the other person stops crying, they continue to cry but are grateful, etc., with the best expected reward in mind)~\cite{johnson2010mental,van2015cognitive}. 

This view of the world model as fundamentally a simulator, rather than a generator of visual content, is gaining broad recognition across the field~\cite{lecun_xing_world_models_2026,li2026functional_taxonomy_world_models}.
While some framings emphasize physical simulation (geometry, physics, and dynamics), we argue that the concept extends further: a world model should serve as the substrate for \textit{simulative reasoning}, the capacity to mentally enact hypothetical scenarios (e.g., physical, social, strategic, or abstract) in service of purposeful decision-making.


Such a WM also enables transfer of knowledge to solving novel tasks, thanks to the fact that real world dynamics, even in different scenarios, share many mechanistic commonalities. For instance, a scuba diver experiences how their body moves in response to low gravity, which is likely helpful to walking on the moon. A mountaineer would be good at predicting how terrain conceals individual movements, which is useful when it's their turn to lead a mountain ambush during a war. On the other hand, a skilled gamer has deep knowledge of how digital characters respond to control signals, which will come in handy when they become drone operators. Therefore, like humans often employ their mental model to help extrapolate from past experiences to act novelly in new environments; machines can leverage a WM similarly to better achieve zero-shot capabilities in unfamiliar environments. 

How should we create such a general WM? The key desiderata for building and training a WM include the following 5 aspects: identifying and preparing training \textbf{data} with the desired world information; adopting a general \textbf{representation} space for the latent world state with possibly richer meaning than the observation data in plain sight; designing an \textbf{architecture} that allows effective reasoning over the representations; choosing an \textbf{objective} that properly guides the model training; determining how to \textbf{use} the world model in a decision-making system. Recent years have seen a surge in efforts toward a world model. In this paper, we examine the empirical design landscape along these five dimensions, survey existing technical approaches including recent systematic proposals, and present our perspective on what a general-purpose world model requires. 

We conclude our examination with a brief preview of an alternative architecture -- PAN -- a \underline{P}hysical, \underline{A}gentic, and \underline{N}ested world model, which we argue offers the potential for a truly general-purpose and actionable world model, based on the following design principles: 1) data from all modalities of experience; 2) mixed continuous and discrete representation; 3) hierarchical generative latent prediction (GLP) modeling paradigm with an extended LLM backbone (for discrete concept-based reasoning), as well as a generative embedding predictive module (for continuous gradient-based reasoning), as the reasoning engines;
4) generative loss grounded in observation data; and 5) usage for simulating experience for training agents with reinforcement learning (RL). Full details of the PAN world models and results will be provided in a separate dedicated manuscript~\cite{xiang2025pan}. 

\section{World Model and Agent Decision-Making}

World model arises in the context of agent decision-making. An agent is an autonomous system that acts in an environment -- the {\it universe}, with both physical and social worlds -- to achieve a goal (e.g., climbing a mountain, winning a military campaign). We consider an environment with discrete timesteps indexed by $t$ (continuous timesteps can be approximated by infinitesimally small discrete time-steps). Formally, the agent takes the current world state $s_t$ and outputs the next action $a_t$ based on a distribution $p_\pi(a_t | s_t)$, known as the ``policy" in reinforcement learning literature. The optimal agent is thus one that best achieves its goals across all environments. The concept of a World Model arises as a surrogate of the environment in general agentic reasoning. 

\subsection{Agent-Environment Model and Optimal Agent}
\label{subsec:optimal-agent}
\begin{figure}
    \centering
    \includegraphics[width=0.85\linewidth]{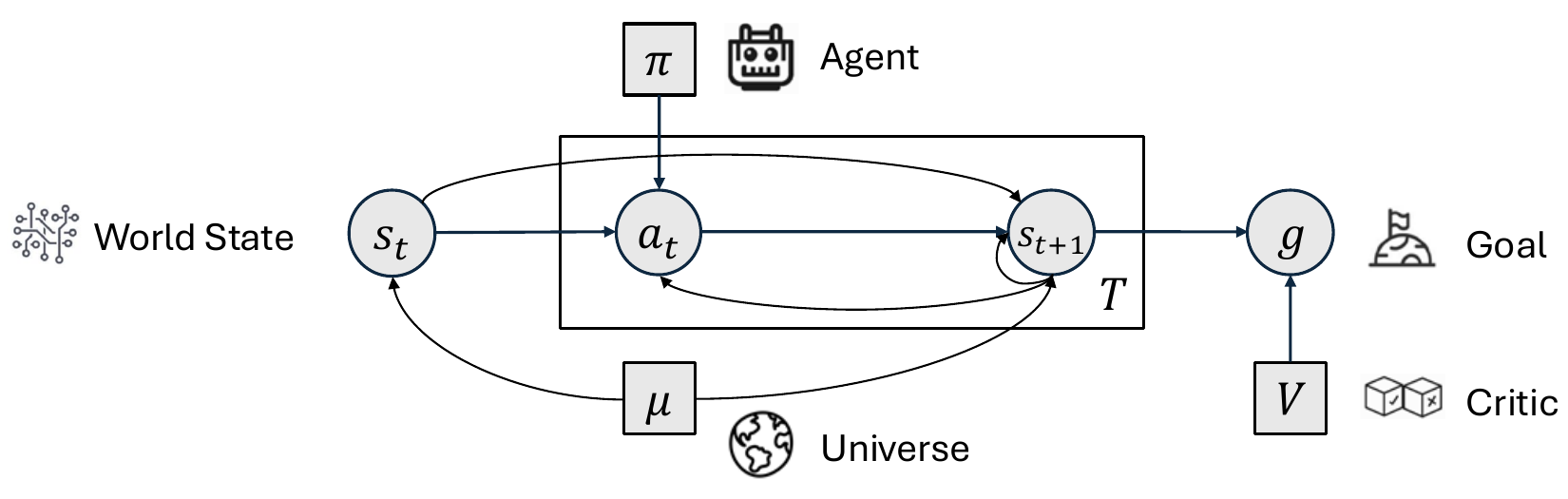}
    \caption{A possible definition of an optimal agent}
    \label{fig:optimal-agent}
\end{figure}

Consider a sequential interaction between an agent and an environment (Figure~\ref{fig:optimal-agent}). At time step $t$, the agent outputs action $a_t$, and the universe $\mu$ takes the current state $s_t$ and the action $a_t$, and outputs the next state $s_{t+1}$ based on distribution $p_\mu(s_{t+1} | s_t, a_t)$. The distribution of the interaction trajectory until timestep $T$, or $(a_t, s_{t+1}, \dots, a_{T-1}, s_T)$, given the current state $s_t$ is thus denoted by:
\begin{align}
  p^{\pi}_{\mu}(a_t, s_{t+1}, \dots, s_T \mid s_t) = \prod_{k=t}^{T-1} {\underbrace {\textstyle p_\pi(a_{k} \mid s_k)}_{\text{ agent }} } \ {\underbrace {\textstyle p_\mu(s_{k+1} \mid s_k, a_{k})}_{\text{ universe }} }  
  \label{eq:trajectory}
\end{align}
In each state $s_t$, the agent also receives a reward $r(g, s_t)$ based on its goal $g$. 
We evaluate the agent by its discounted cumulative reward, denoted as $\sum_{k=t}^\infty \gamma_k r(g, s_k)$ (with the discount parameter $\gamma_t$ decaying to zero with time, i.e., $\lim_{t \rightarrow \infty} \gamma_t = 0$).
Note that this reward function can be dense (e.g., gaming scores), but perhaps frequently sparse (e.g., curing a disease).
The agent's long-term success can thus be measured by its expected future discounted reward, also known as \textbf{value function}~\cite{sutton1998reinforcement}: 
\begin{align}
    V_{\pi,\mu}^g(s_t) &\vcentcolon= \mathbb{E}_{\pi,\mu} \left[ \sum_{k=t}^\infty \gamma_k r(g, s_k) \mathrel{\bigg|} s_t \right] \nonumber \\
    &= \lim_{T \rightarrow \infty} \sum_{(a_t, s_{t+1},\dots, s_T)} {\underbrace {\textstyle \sum_{k=t}^T \gamma_k r(g, s_k) }_{ \text{goal} } } \ {\underbrace {\textstyle p^{\pi}_{\mu}(a_t, s_{t+1}, \dots, s_T \mid s_t)}_{ \text{trajectory} } }
    \label{eq:value-function}
\end{align}
Based on Equations~\ref{eq:trajectory} and \ref{eq:value-function}, we can define the optimal agent in this universe $\mu$ as one that maximizes the value function, written formally as below:
\begin{equation}
    \pi^*_\mu := \argmax_\pi V^{g}_{\pi,\mu}
\end{equation} 
Some simple derivation will show that the optimal agent in state $s_t$ will select actions based on the following decision rule $\pi^*_\mu(s_t)$ when planning for actions $a_{t:T-1}$:
\begin{equation}
    \pi^*_\mu(s_t) = {\underbrace {\argmax_{ a_{t:T-1} } }_{\text{possible actions} } } \ \sum_{s_{t+1:T}}   \Bigg({\underbrace { \sum_{k=t}^{T-1} \gamma_k r(g, s_k) + \gamma_T V_{\pi, \mu}^g(s_T) }_{\text{goal progress}} }\Bigg) \prod_{i=t}^{T-1} \ {\underbrace {p_\mu(s_{i+1} | s_i, a_i)}_{\text{universe response}} } \
    \label{eq:optimal-decision-making-mu} 
\end{equation}

\subsection{World Model and Simulative Reasoning}

\label{subsec:world-model}

\begin{figure}
    \centering
    \includegraphics[width=0.85\linewidth,page=2]{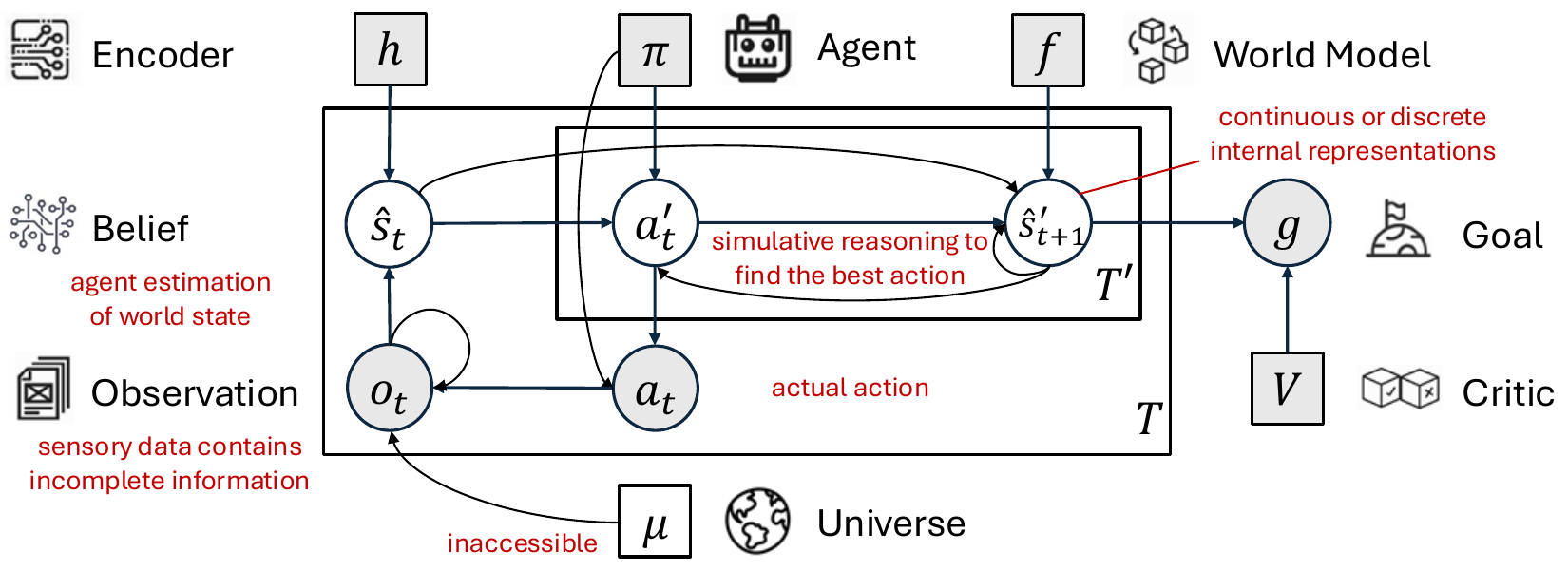}
    \caption{An agent in real world where groundtruth world state and universe are unavailable to experience or experiment, so world model is crucial for simulation.}
    \label{fig:intro-world-model}
\end{figure}

Note that optimal decision-making defined in Equation~\ref{eq:optimal-decision-making-mu} requires the agent to have access to the ground-truth world state $s$ from the universe $\mu$ to experience and optimize. However, these are often not the case aside from simple scenarios like the Go and Chess games~\citep{silver2016mastering,silver2017mastering} -- imagine building an agent to land on Mars, or even a real-world robot relying on noisy sensors in daily environments. World Model thus arises as a crucial component for predicting the universe's response to a general agent. 
Specifically, as illustrated in Figure~\ref{fig:intro-world-model}, a WM $f$ operates on an internal (continuous or discrete) representation of the world state, denoted as a \textit{belief state} $\hat{s}_t$, which is derived from sensory inputs $o_t$ via an \textit{Encoder} $h$ (unlike the optimal agent described in \S\ref{subsec:optimal-agent} which has direct access to the true world state $s_t$).
Given a proposed action $a'_t$ (as opposed to the true action $a_t$ used by the optimal agent), the WM predicts the next belief state $\hat{s}_{t+1}$ according to the distribution $p_f(\hat{s}_{t+1} | \hat{s}_t, a'_t)$. 
Note that for such simulative reasoning to be possible at all, the belief state must be \textit{stateful}: $\hat{s}_t$ must constitute an identifiable, persistent estimate of the world state that the agent can hold in long-/short-term memory, revisit, and update across simulation steps, rather than a transient encoding of the current sensory stream. We return to this requirement, and to what representations satisfy it, in \S\ref{subsec:critique-representation}.
This predicted belief state then allows the agent to propose the next action, continuing the cycle of prediction and action up to a desired time horizon $T'$.

The agent can simulate multiple such sequences of proposed actions and belief states, and select the actual action $a_t$ (upon observing $o_t$) based on some external function such as a \textit{Critic} that evaluates the outcome against a given \textit{Goal}.
Thus, a WM essentially functions as a generative model of possible future world states, which enables \underline{simulative reasoning}, or "thought experiments". 
Formally, for the optimal agent $\pi^*_f$ equipped with WM $f$ in belief state $\hat{s}_t$, we define
the simulation-based decision rule in Equation~\ref{eq:world-model-decision-making} as follows:
\begin{equation}
    \pi^*_f(\hat{s}_t)= {\underbrace {\argmax_{ a'_{t:T'-1} } }_{\text{possible actions} } } \ \sum_{\hat{s}_{t+1:T'}} \Bigg({\underbrace { \sum_{k=t}^{T'-1} \gamma_k r(g, \hat{s}_k) + \gamma_{T'}V_{\pi, f}^g(\hat{s}_{T'}) }_{\text{goal progress}} }\Bigg)  \prod_{i=t}^{T'-1} \ {\underbrace {p_f(\hat{s}_{i+1} | \hat{s}_i, a'_i)}_{{\scriptsize \shortstack{simulation with\\world model}}}}
    \label{eq:world-model-decision-making}
\end{equation}

A general-purpose WM enables simulation of diverse possibilities across a wide range of domains, enabling agents to reason about outcomes without direct interaction with the environment. This includes, but is not limited to the following examples:
\begin{itemize}
    \item \textbf{Physical dynamics}: Mechanics of the real world, such as how water pours, how an object moves when thrown, or how a machine operates under varying conditions.
    \item \textbf{Embodied experiences}: Internal bodily states (e.g., balance, posture), sensations (e.g., heat, pain, dizziness), and complex motor activities like getting dressed or tying shoes.
    \item \textbf{Emotional states}: Affective responses such as happiness, sadness, or fear, which can facilitate planning in emotionally charged contexts (e.g., therapy or social interactions).
    \item \textbf{Social situations}: The actions and internal states of other individuals, including their embodied or emotional experiences, needs, intentions, and expectations. 
    \item \textbf{Mental world}: Abstract ``thought processes'' such as logistics, tactics, and strategies, potentially in multi-agent or adversarial settings.
    \item \textbf{Counterfactual world}: Alternative realities or ``what if'' scenarios to guide better decision-making under uncertainty or incomplete information.
    \item \textbf{Evolutionary world}: Generational dynamics such as genetic inheritance, adaptation, and survival of organisms.
\end{itemize}
As stated earlier, a major function of the WM is to enable {\it simulative reasoning} -- where an agent performs a series of thought experiments by simulating the outcomes of plans with the WM, based on which it can choose the best plan. This reasoning approach contrasts with alternative approaches also explored in AI systems, such as {\it logical reasoning} inspired by a type of human mental activity that aims to arrive at a conclusion through rigorous inferences or arguments starting from a set of premises and then using stepwise relational formal reasoning (e.g., Lambda calculus) to reach a conclusion supported by these premises; or {\it model predictive control} based reasoning used in the process industries in chemical plants and oil refineries to infer optimal action sequence through mathematical programming (e.g., convex optimization) to satisfy a set of constraints. A hallmark of 
simulative reasoning enabled by a WM is its flexibility, generalizability, and scalability with respect to changing computing resource, memory, environment, and problem complexity, thanks to the WM's intrinsic ability to simulate all possibilities across domains. In this regard, a WM bears important empirical (e.g., end-to-end experience) similarity to a modern LLM such as ChatGPT that can operate in a subject-agonistic way in the lingual intelligence space. Indeed, as we discuss later, a general-purpose WM can make use of LLMs as its key building blocks. 

Combined with an encoder that estimates beliefs of the world states from arbitrary sensory observations, WM supports machines to perform thought experiments computationally with controlled depth (i.e., number of steps) and width (number of trajectories).  AlphaGo~\citep{silver2016mastering}, for example, can be seen as a special case of simulative reasoning with Monte-Carlo Tree Search (MCTS) using a known (trivial) WM. In the physical world, simulative reasoning enables autonomous vehicles to drive safely by forecasting future street scenarios~\citep[e.g.,][]{varadarajan2022multipath++,sun2022m2i} or military commanders to develop battle-winning tactics by anticipating the outcomes of troop movements~\citep{strauch1982battle}.
WM also enables simulations on different time scales, allowing one to answer questions about billions of years on Earth's evolutionary path, or just a few moments on a hypothetical Martian civilization. 

\section{The World Model Landscape}

Recent work on world modeling have led to a variety of systems, many of which optimized for a specific domain or type of simulation. Interestingly, a commonality nevertheless can be found in these diverse systems in that they all put a significant emphasis on video/image generation, and on visual quality of the generated contents:

\paragraph{Gaming World Models} 
Systems such as Genie 2~\cite{parkerholder2024genie2} (Google DeepMind), Muse~\cite{kanervisto2025world} (Microsoft), and Oasis~\cite{oasis2024} (Decart and Etched) simulate video game environments using generative models. These models can render plausible trajectories from visual and action input, producing up to 1-2 minutes of continuous gameplay content. 
More recently, Genie 3~\cite{genie3} and Ant Group's LingBot-World~\cite{team2026advancing} simulate highly realistic 3D environments with navigation-based actions and promptable world events.
Despite their advances, these systems remain domain-specific -- for instance, Genie 2 and Muse rely on restrictive console-style inputs, while Oasis is limited to Minecraft-like settings -- and action control remains limited, with the inclusion of interacting agents still largely absent.
Furthermore, their temporal coherence remains shallow, as current generation horizons (1-2 minutes) fall short of representing full gameplay sessions, which often span several hours.
As such, gaming world models lack the flexibility, generality, and long-term reasoning capabilities required for more open-ended, agent-driven tasks.



\paragraph{3D Scene World Models} World Labs~\cite{worldlabs2025generating} and related efforts focus on stylized 3D scene generation and egocentric navigation. More recently, World Labs released additional details about its system Marble~\cite{worldlabs2025marble}, which generates explorable 3D environments from image or text prompts, and Meta published WorldGen~\cite{wang2025worldgen} with similar functionalities. While visually appealing and demonstrating strong spatial realism, there is currently no clear indication that these models support interactions beyond simple navigation. Richer simulation capabilities appear to rely on exported Gaussian-splat representations, which operate outside the world model itself. 
This amounts to a \textit{program-as-simulator} approach: the learned model generates geometric structure, but the dynamics (e.g., how objects move, collide, and deform) are governed by prescribed physics engines, making it closer to a digital twin than a learned world model. 
Indeed, from available demonstrations, these models simulate static environments without dynamic agents, physics, or rich interactivity. 
This results in incomplete simulations that are insufficient for tasks involving physical causality, multi-agent behavior, or goal-driven planning. Although these systems push the boundary of spatial realism, they do not support full-fledged world modeling for decision-making or agent learning.

\paragraph{Physical World Models} Generative models such as Wayve GAIA-2~\cite{russell2025gaia} and NVIDIA Cosmos~\cite{agarwal2025cosmos} are trained specifically on physical control tasks, including autonomous driving, robotic manipulation, and embodied navigation. These systems demonstrate impressive fidelity in modeling low-level physics and sensory-motor control under diverse conditions (e.g., varying weather, lighting, and geography). More recently, NVIDIA has introduced Cosmos-Predict2.5~\cite{ali2025world} and Cosmos 3~\cite{nvidia2026cosmos3}, while 1X developed 1XWM~\cite{ho20251x}, further improving fidelity and control in autonomous driving and robotic manipulation. Runway ML has also introduced GWM-1~\cite{runway2025gwm1}, described as a ``general world model family,'' though based on currently available information it functions more as a collection of specialized models targeting robotics, 3D environments, and avatar-based scenarios. However, these systems remain tightly coupled to their respective domains, often relying on task-specific sensors, data, or control architectures. 
They excel in their respective constrained settings, but have yet to address the broader challenge of simulating complex, multi-agent, or socially grounded worlds.

\paragraph{Video Generation Models} 
Another popular class of models focus on general-purpose video generation, with recent examples including OpenAI's Sora~\cite{brooks2024video}, Google DeepMind's Veo~\cite{deepmind2025veo}, and ByteDance's Seedance 2.0~\cite{bytedance_seedance2}.
These models aim to generate high-quality video sequences from textual prompts and/or prior frames, with the latest systems also supporting multimodal outputs and interactive extension or editing capabilities. While visually stunning, these models only generate fixed trajectories and do not support interactions based on alternative actions. Specifically, they lack explicit notions of state, action, and possibly even object-level or conceptual representation within the video frames. They also provide no simulation control that would allow for reasoning about counterfactual outcomes or evaluate different decisions. Consequently, these systems fall outside the definition of world models for reasoning and planning, and are better understood as content-generation tools (focused on pixel-level synthesis) rather than parts of decision-making systems.


\paragraph{Joint Embedding Predictive Models} 
Last but not least, the joint embedding predictive architecture (JEPA) family, including the V-JEPA series~\cite{bardes2024revisiting,assran2025vjepa2selfsupervisedvideo}, DINO-WM~\cite{zhou2024dino}, and PLDM~\cite{sobal2025learning} from Meta FAIR, has attracted significant attention for its conceptually elegant approach to world modeling.
These models forego pixel-level generation and instead predict future latent embeddings, often using an encoder-encoder architecture and supervising the outputs in the latent space with energy-based losses. While this design promises to improve tractability, the evidence for practical usability remains scarce as these models have mainly been demonstrated in toy environments with simple heuristics and action spaces. The very recent V-JEPA 2~\cite{assran2025vjepa2selfsupervisedvideo} marks a step forward by applying joint embedding prediction to robotic arm manipulation tasks, but it remains unclear whether such models can generalize across more diverse tasks (e.g., making breakfast) or scale to higher-complexity environments with long-term dependencies (e.g., mountaineering).

In summary, although the systems surveyed above have demonstrated significant progress in modeling aspects of the world, most fall short of enabling purposeful reasoning and planning in real world applications due to limitations in scope, abstraction, 
controllability, interactability, and generalizability. 
In particular, except for JEPA, contemporary WM systems almost universally emphasize video generation as a core function, yet this emphasis remains largely unexamined and lacks a compelling justification.
This focus may reflect the underlying conceptual ambiguities, or even misunderstandings, regarding a fundamental question -- what \textit{is} a world model?
We argue that {\bf a world model is \textit{not} fundamentally about generating videos, but about serving as a simulator for reasoning and thought-experiment}, which was echoed in recent discussions~\cite{lecun_xing_world_models_2026,li2026functional_taxonomy_world_models}.
Our discussion below is organized around this definition to examine the plausibility and feasibility of current technical approaches to world modeling. 

\section{Critique of World Modeling}

Building a general-purpose world model capable of supporting simulative reasoning across diverse domains requires addressing several interrelated design dimensions: what \textbf{data} to train on, how to \textbf{represent} world states, what \textbf{architecture} to use for prediction, what learning \textbf{objective} to optimize, and how the trained model is \textbf{used} for decision-making. Across the diverse efforts surveyed above, several common wisdoms regarding these dimensions have emerged, which can be summarized as follows:
\begin{enumerate}
    \item \textbf{Sensory inputs should be emphasized over text}, because of the larger data volume from the physical world (e.g., a 4-year-old processes approximately $1.1^{14}$ bytes of visual data, whereas all textual data used to train modern LLMs amount to approximately $0.9^{14}$ bytes).
    \item \textbf{World states should be represented as continuous vectors} to enable gradient-based optimization, rather than discrete tokens.
    \item \textbf{An encoder-encoder architecture that predicts the next latent state} is preferred over autoregressive generation of obervations, to avoid compounding prediction errors.
    \item \textbf{Supervision by latent reconstruction objectives} is preferable to reconstructing raw observations, due to not modeling irrelevant or unpredictable details.
    \item \textbf{The world model should be used for action selection via model-predictive control (MPC)} rather than reinforcement learning (RL), for sample efficiency and safety.
\end{enumerate}
\begin{figure}
    \centering
    \includegraphics[width=0.7\linewidth]{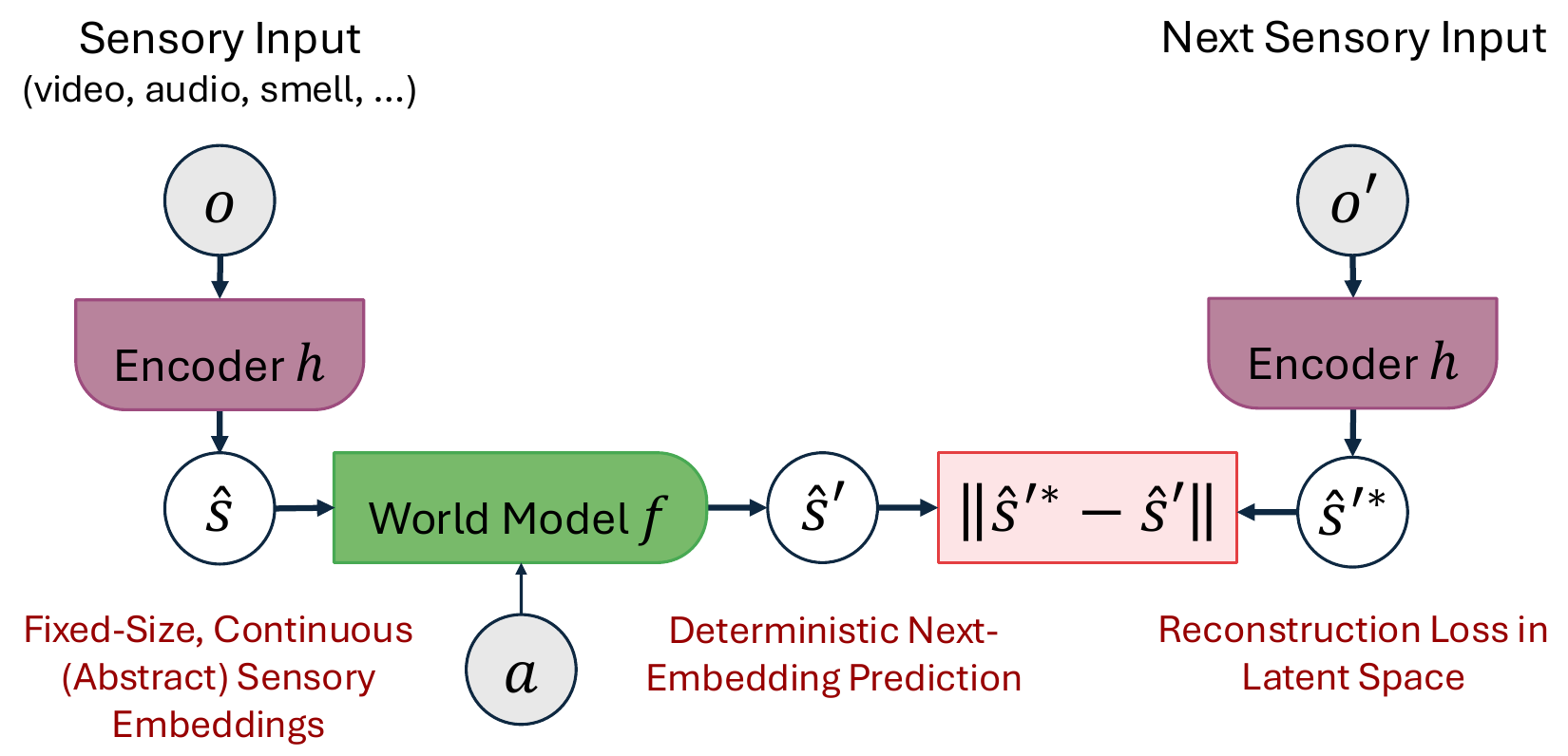}
    \caption{A systematic technical proposal that results from common wisdom on world modeling, most prominently instantiated by the joint embedding predictive architecture (JEPA).}
    \label{fig:critique-framework}
\end{figure}
These common wisdoms have led to a systematic set of technical proposals, most prominently instantiated in the joint embedding predictive architecture (JEPA)~\cite{lecun2022path}. As illustrated in Figure~\ref{fig:critique-framework}, this line of work is centered around an idea that can be summarized as ``next representation prediction'' rather than next data prediction: 
\begin{itemize}
    \item \textbf{Sensory-First Pretraining}: The framework prioritizes continuous sensory data such as video, audio, and other modalities over text.
    \item \textbf{Fixed-Size, Continuous State Embeddings}: Given sensory input $o$, an encoder $h$ estimates the world state $\hat{s} = h(o)$ as an abstract continuous embedding with fixed dimensions (e.g., $\hat{s} \in \mathbb{R}^d$). 
    \item \textbf{Encoder-Encoder Architecture}: Based on action input $a$, the world model $f$ predicts the next state embedding $\hat{s}' = f(\hat{s}, a)$ in a deterministic manner. In particular, the architecture does not use a decoder $g$ to reconstruct the next observation $o'$, but instead applies the encoder again to bootstrap $\hat{s}'^* = h(o')$ as a \textit{ground-truth next state} for supervision.
    \item \textbf{Reconstruction Loss in Latent Space}: Instead of supervising the world model by the difference between the reconstructed next sensory input $\hat{o}'$ and the actual data $o'$, this framework bases learning on the deviation between the predicted next state $\hat{s}'$ and the bootstrapped ground truth $\hat{s}'^*$ (e.g., L2 loss $\lVert \hat{s}' - \hat{s}'^* \rVert$).
    \item \textbf{Action Selection via MPC}: Given current observation $o_t$, the framework favors proposing an initial action sequence $(a_t, a_{t+1}, \dots, a_{T-1})$, using the world model $f$ to simulate the next states $(s_{t+1}, s_{t+2}, ..., s_T)$, and optimizing the actions based on goal progress $V_g(s_T)$.
\end{itemize}

These design choices represent an intellectually coherent perspective on world modeling and make important contributions -- particularly the emphasis on learning abstract, predictable structure rather than reconstructing every low-level detail, and the recognition that a world model should focus on what is learnable and relevant~\cite{lecun_xing_world_models_2026}. 
However, when the goal is a general-purpose world model capable of supporting simulative reasoning across diverse domains, each of these choices involves tradeoffs that merit careful examination. In the following, we analyze each design dimension, survey the landscape of existing approaches, and present our perspective on what a general-purpose world model requires.

\subsection{Data: Information Density, Not Just Volume}
\label{subsec:critique-data}


What data should a world model be trained on? One perspective emphasizes sensory inputs due to their larger raw volume compared to text: as estimated, LLMs are trained on $0.9 \times 10^{14}$ bytes of textual data, whereas a 4-year-old child has already seen $1.1 \times 10^{14}$ bytes of vision data~\cite{lecun2022path}.

Although sensory data streams such as video appear massive in raw volume, much of that data is low in semantic content and highly redundant~\citep{gonzales1987digital,wang2002video}. In contrast, natural language is an evolved compression of human experiences, optimized over generations of abstract communication and conceptual reasoning~\citep{mackay2003information,pinker2007language}.

Text captures not only physical realities but also mental, social, and counterfactual phenomena, which are otherwise difficult or impossible to observe directly~\citep{deacon1998symbolic}. For example, concepts such as justice, motivation, or regret are richly encoded in language but have no direct sensory equivalent. Moreover, language provides an interface to collective human memory (e.g., documented observations, scientific discoveries, engineering failures), which are difficult, if not impossible, to derive from raw perceptual input alone~\citep{vygotsky2012thought}. Indeed, models trained on text can write software~\citep{wang2024openhands} or solve Olympiad-level math problems~\citep{chervonyi2025gold}, whereas those trained on raw visual and motion data alone have been more suited for physical navigation~\citep{waymo2025driver} or manipulation tasks~\citep{figure2025helix}.

Thus, the path towards general-purpose world modeling must leverage all modalities of experience, whether it be text, images, videos, touch, audio, or more. 
Crucially, these modalities are not interchangeable, but reflect different layers of experience (e.g., video captures spatiotemporal dynamics in the embodied and physical world, while language encodes abstract concepts and social norms). A general-purpose world model should not favor any single modality at the expense of others, but rather learn from this stratified structure of experience to generalize across diverse tasks. Ignoring any level, be it low-level perception or high-level abstraction, risks omitting crucial information needed for intelligent behavior.

\subsection{Representation: Continuous? Discrete? Or Both?}
\label{subsec:critique-representation}




How should world states be represented? Current approaches (e.g., V-JEPA 2~\cite{assran2025vjepa2selfsupervisedvideo} and Cosmos 3~\cite{nvidia2026cosmos3}) typically show a preference for continuous embeddings over discrete tokens, arguing that continuous representations enable smoother gradient-based optimization and richer encoding of perceptual detail.

Do humans perform gradient optimization (e.g., SGD) over continuous neural signals or pattern search (e.g., kNN) over discrete concepts during reasoning? We don't know for sure. What we do know is that reasoning can be cognitive or physiological, or both, and it might be unlikely to have one algorithm fit for all. 

\begin{figure}[t]
    \centering
    \includegraphics[width=1.0\linewidth]{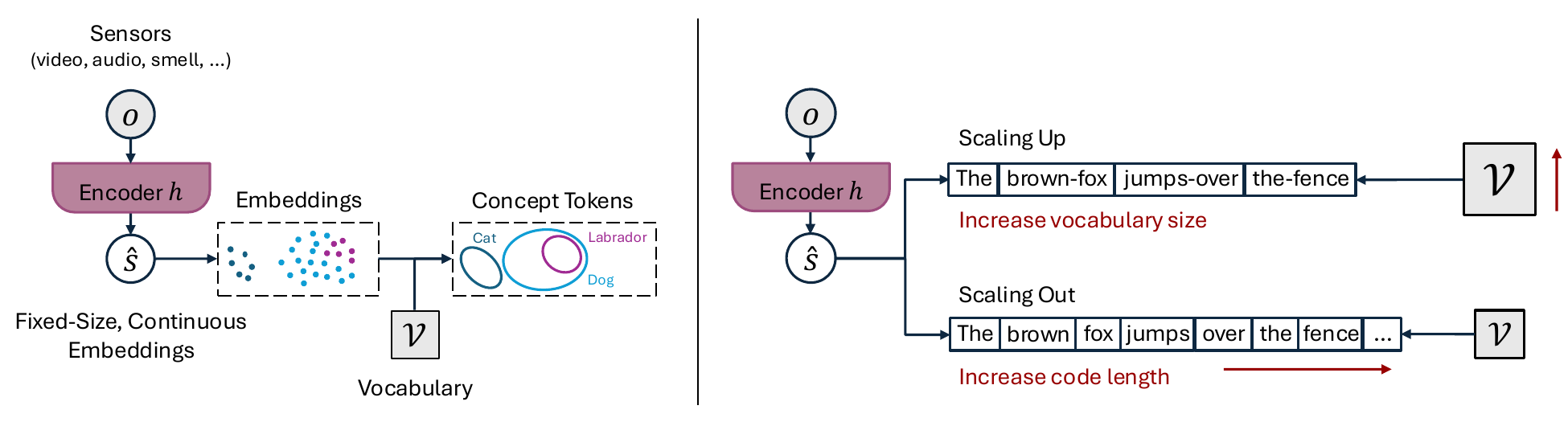}
    \caption{Vocabulary-based tokens is an effective way to categorize perceptual inputs into discrete concepts for reasoning (left). We may scale up or scale out discrete code to deal with increasing data complexity (right). Thm.\ref{thm:completeness-of-language} shows either is effective, but scaling out is more efficient.}
    \label{fig:enter-label}
\end{figure}

While continuous representations allow for smoother gradient flow, an important complementary consideration is the inherent noise and high variability associated with continuous sensory inputs, which can make them brittle for certain forms of reasoning. 
In effect, fixed-size continuous embeddings derived directly from raw sensory streams are \textbf{stateless}: they fluctuate with noise, viewpoint, and other nuisance factors rather than with the underlying world, and therefore have difficulty with functioning as identifiable states that can anchor reasoning across time.
Human cognition has evolved to counter this variability by categorizing raw perception into \textit{discrete concepts}~\citep{barrett2017emotions}, which are what we typically encode in language, symbols, and structured thoughts (Figure~\ref{fig:enter-label}, left).

By contrast, we call a representation \textbf{stateful} if it constitutes an identifiable, persistent, and semantic estimate of the world state -- one that can be stored in, recalled from, and reasoned over as long- and short-term memory, that remains stable under nuisance variation in the sensory stream, and that is sufficient for predicting the dynamics that matter (i.e., a genuine \textit{state} in the state-space sense of the belief state $\hat{s}$ introduced in \S\ref{subsec:world-model}). For world modeling, we argue, statefulness is not optional but a must: a representation that cannot be re-identified and carried forward in memory cannot support simulative reasoning over extended horizons.

Vocabulary-based tokens are thus not a liability, but an asset: they offer a stable, composable medium for representing concepts at various levels of abstraction. They form the foundation for designing and building language-based AI systems of today, such as the LLMs, which ground reasoning on 
sequence of discrete words which are human tokens that correspond to diverse perceptions from the universe (e.g., physical, mental, or social worlds), and allow a form of long-term memory to be employed by implementing (ideally, dynamically controllable) context length.  Although not entirely accurate, it is reasonable to consider the space of language as a man-made (through evolution and learning) \textit{latent space} of the perceived and describable universe that humans live -- a substantial subspace of the whole universe. 
Benefitting from massive text-based pretraining, an LLM can learn to simulate contents in this latent space formed by natural language. Indeed, recent work that represents world states in natural language has seen success for reasoning and planning in a wide range of practical tasks~\cite{hao2023reasoning,simura2025}. 
Complementing natural language tokens, modern techniques like VQ-VAE~\cite{van2017neural} allow us to further convert sensory data (e.g., images or audio) into discrete tokens while preserving structure and semantics.

While such discrete representations are expected to offer stability and symbolic structure like natural language tokens, a natural concern is whether they can faithfully capture the richness of high-dimensional, continuous sensory data due to the risk of information loss through distillation. This concern grows with the complexity of the world: Will discrete tokens be sufficient for distinguishing between subtly different world states?
Indeed, the world often contains deeper layers of meaning than what is directly observable through sensory input (e.g., a puppet's movements may reflect the hidden intentions of the puppet master). Capturing such latent structure requires representations that can scale in expressive capacity.

In an attempt to understand more deeply the potential and limitation of using discrete tokens, we present a theoretical result showing that discrete representations can, in principle, preserve arbitrarily fine distinctions between real-valued inputs, provided we scale them appropriately. Specifically, we consider two intuitive strategies for increasing representational capacity:
\begin{itemize}
    \item \textbf{Learning a larger modality tokenizer} (scale up): Keep the number of tokens fixed, increase the vocabulary size to allows each token to encode a finer-grained chunk of information.
    \item \textbf{Finding a longer language expression} (scale out): Keep the vocabulary fixed, increase the sequence length and combine more tokens to express more complex inputs. 
\end{itemize}
As we show in Theorem~\ref{thm:completeness-of-language} below, it is more efficient to scale out by increasing the length of the encoding.

\begin{theorem}[Completeness of Language Representation]
    \label{thm:completeness-of-language}
    Assume real inputs $\mathbf{x} = [x_1, \dots, x_T]$, where $x_t \in \mathbb{R}^D$ and $\| x_t \| < K$. For any $\epsilon > 0$, there exists a language $L_\epsilon = (\mathcal{V}, N, f_\epsilon)$ with vocabulary $\mathcal{V}$, maximal sentence length $N < \infty$, and a mapping function $f_\epsilon: \mathbb{R}^{TD} \rightarrow \mathcal{V}^N$ such that for all $\mathbf{x}, \mathbf{x'} \in \mathbb{R}^{TD}$, $\| \mathbf{x} - \mathbf{x'} \| > \epsilon \Rightarrow f_\epsilon(\mathbf{x}) \neq f_\epsilon(\mathbf{x}')$.
\end{theorem}
\begin{explanation}
    If you have a sequence of continuous sensor readings or data points, no matter how small a difference you want to be able to distinguish between two sequences, you can always create a language (a system of words or symbols) that can represent these sequences uniquely.
\end{explanation}
\begin{proof}[Proof Sketch]
We will prove the contrapositive that $f_\epsilon(\mathbf{x}) = f_\epsilon(\mathbf{x}') \Rightarrow \| \mathbf{x} - \mathbf{x}' \| \leq \epsilon$. Specifically, we propose two ways to scale the discrete code: 
\begin{itemize}
    \item \textbf{Case 1} (Learning a Larger Modality Tokenizer). Keep the code length constant at $T$, increase the vocabulary size to $M_\epsilon = \lceil \sqrt{TD} \tilde{K} \tilde{\epsilon}^{-1} \rceil ^D$ (i.e., \textbf{scaling up}).
    \item \textbf{Case 2} (Finding a Longer Language Expression). Keep the vocabulary size constant at $M$, increase the maximum sentence length to $N_\epsilon = TD \lceil \log_M \sqrt{TD} \tilde{K} \tilde{\epsilon}^{-1} \rceil$ (i.e., \textbf{scaling out}).
\end{itemize}
\end{proof}
The detailed proof can be found in Appendix~\ref{appendix:proof-completeness-of-language}. As the proof shows, complete representations are achievable with vocabulary-based discrete tokens. However, the way to scale the representation matters (Figure~\ref{fig:enter-label}, right). In Case 1, the vocabulary size must grow in $\mathcal{O}((TD)^D)$, i.e., exponentially with the input size we'd like to capture, which is likely not sustainable. In Case 2, on the other hand, the sequence length need only increase in the order of $\mathcal{O}(TD \log TD)$, which is much more manageable. So in theory, scaling out the token sequences, which can be implemented with an enhanced LLM (with visual tokenizers and vocabulary mergers or switchers), offers a more flexible and efficient pathway to capturing complex structure in data generally. 
In other words, opting for discrete, vocabulary-grounded codes for the sake of statefulness need not come at the cost of expressive fidelity: Theorem~\ref{thm:completeness-of-language} guarantees that suitably scaled discrete codes can preserve arbitrarily fine distinctions between world states, answering the information-loss concern raised above.
In practice, the model may also represent complex inputs efficiently by dynamically resizing the dictionary, and describe novel inputs by growing the vocabulary with more new observations. 

In summary, given that discrete and continuous latent representations offer complementary level of abstraction, representation power, and operationalizability, we advocate for the approach of \textit{mixed representations},
in which discrete tokens supply the \textit{stateful} backbone of the world state -- a stable, identifiable, and memory-anchored medium
for more robust, interpretable, and symbolic forms of reasoning, while continuous embeddings still play a role in capturing fine-grained sensory nuance. While this form of mixed representation is still in its early stages, recent work has demonstrated its promise for generalization in world modeling~\citep{xiang2024pandora} and other forms of reasoning~\citep[e.g.,][]{mitra2024compositional}.

\subsection{Architecture: Pure Abstraction, or Autoregressive Generation}
\label{subsec:critique-architecture}


What architecture should a world model adopt? Approaches range from direct next-observation generation to joint embedding prediction in latent space, each involving distinct tradeoffs in grounding, efficiency, and scalability. In this subsection, we analyze two competing approaches, examine their tradeoffs for general-purpose world modeling, and present our blueprint of a generative latent prediction (GLP) architecture as the synthesis.

\subsubsection{Two Competing Approaches}

The first approach, exemplified by encoder-decoder-based systems such as Genie 2~\cite{parkerholder2024genie2} and NVIDIA Cosmos series~\cite{agarwal2025cosmos,ali2025world,nvidia2026cosmos3}, prioritizes \emph{validation} by training video-generation architectures that directly predict the next observation $o'$ from the current observation $o$ and action $a$, resulting in a \emph{closed loop} that checks predictions against observable data.


An influential alternative, represented by the JEPA framework~\cite{lecun2022path}, prioritizes \emph{abstraction} through a non-autoregressive, non-generative, encoder-encoder design that predicts the next latent state directly, sidestepping the need to reconstruct raw observations in an \emph{open loop} setup. 
This represents a deliberate and intellectually coherent design choice, motivated by the valid concern that autoregressive observation generation may compound errors over long sequences and waste capacity on irrelevant details. 
As we analyze below, however, 
the recursive application of latent prediction is itself functionally autoregressive and generative, and the practical limitations attributed to autoregressive models may be addressed through architectural choices rather than architectural avoidance. 

\begin{figure}[t]
    \centering
    \includegraphics[width=1\linewidth,page=2]{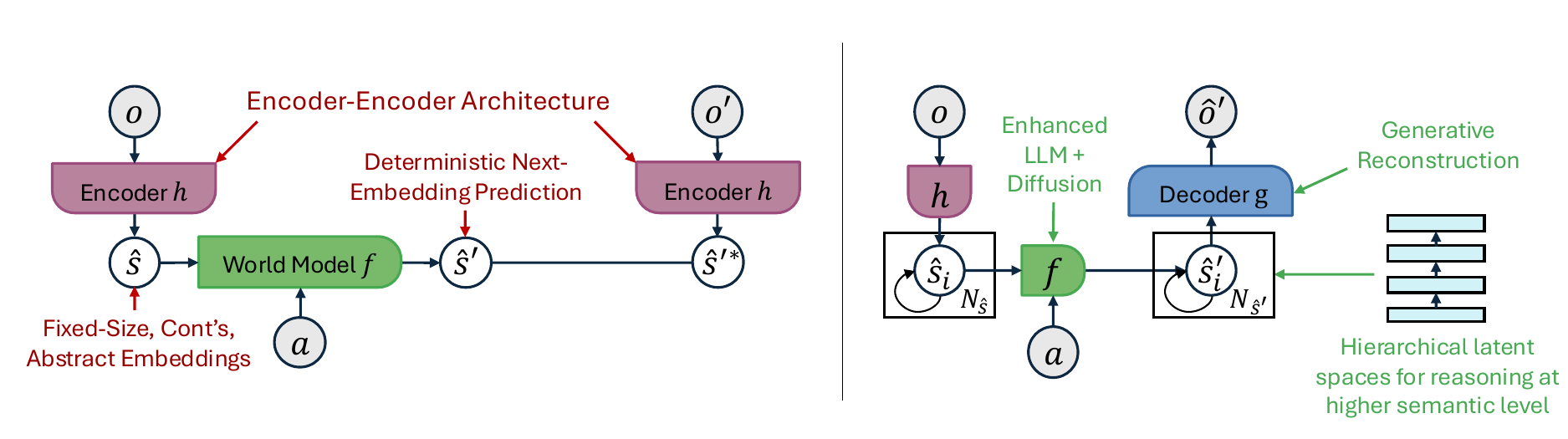}
    \caption{Comparison of the joint embedding predictive architecture (JEPA, left) and the proposed Generative Latent Prediction (GLP) architecture (right). JEPA operates in open loop, predicting the next latent state without reconstructing observations. GLP closes the loop through a decoder that validates predictions against observable data, while incorporating both an enhanced LLM backbone for discrete reasoning and a diffusion-based predictor for continuous dynamics.}
    \label{fig:critique-arch}
\end{figure}

Below, we provide a systematic analysis of JEPA and an alternative architecture we propose that not only inherits important advantages (e.g., abstraction) of JEPA, but also draws inspiration from encoder-decoder approaches (e.g., validation) such as Genie. 
Formally, JEPA defines two core functions (Figure~\ref{fig:critique-arch}, left): 
$$\hat{s} = h(o), \quad \hat{s}' = f(\hat{s}, a),$$
where $h$ is an encoder from observations to latent states, and $f$ is the world model which predicts the next latent state given the current state and an action. Recursive application of these two operators defines a latent transition model that is effectively autoregressive and generative, even though it symbolically lacks an explicit probabilistic decoder to generate what can be compared against real next observation data. (It does not mean such comparisons are avoided, as the second encoder at the output end, in fact, indirectly still makes that comparison, but with poorer mathematical controllability, as we discuss in the next section.) More precisely, JEPA can be viewed as specifying a degenerate conditional distribution, denoted informally as below: 
$$p_f(\hat{s}' | \hat{s}, a) = \delta(\hat{s}' - f(\hat{s}, a)),$$
where $\delta(\cdot)$ is the Dirac delta function centered at the deterministic prediction. Thus, JEPA is not generative in the \textbf{probabilistic} sense (i.e., it does not model uncertainty or samples from a distribution over outcomes), but it is generative in the \textbf{functional} sense of recursively simulating the evolution of the latent states over time, and is therefore subject to the same issues of autoregressive models. 
This is not to say, however, that autoregressive models are inherently flawed due to error accumulation. Many real-world systems (e.g., three-body problem, fluid dynamics, or financial markets) are fundamentally chaotic, with small deviations growing exponentially over time~\citep{perko2013differential}. In such settings, exact prediction is impossible regardless of model class. However, well-structured autoregressive models (e.g., Kalman filters~\cite{baum1966statistical} for continuous cases and HMMs~\cite{baum1966statistical} for discrete cases) can still learn useful, abstract properties of the system (e.g., whether water will spill, the direction of price movement) that are often surprisingly stable and predictable -- an insight grounded in ergodic theory and statistical mechanics~\citep{ornstein1991statistical}.

\subsubsection{Tradeoffs for General-Purpose Modeling}

A common concern with encoder-decoder architectures (which defines an additional function $\hat{o}' = g(\hat{s}')$ with $g$ as a decoder from latent states back to observations) is that they may compel the model to reconstruct aspects of the environment that are either inherently unpredictable or irrelevant to task performance. 
Examples often cited include fine-grained visual details, inconsequential events, or out-of-scene content, which may mislead the model into learning unstable or spurious correlations.
Proponents of encoder-only architectures thus suggest that by avoiding this reconstruction step, the resulting WM can focus more selectively on predictable and task-relevant elements.  

However, the abstraction-first approach introduces a complementary risk: a predictor can only be as good as the representation it operates on.
In JEPA, the encoder $h$ is intentionally lossy and non-invertible, discarding information deemed irrelevant. But if the encoder discards information that turns out to matter for the dynamics being predicted, the world model is learning transitions in an ill-formed state space, and no amount of predictor sophistication can recover what was thrown away.
Put differently, a lossy, non-invertible encoder cannot yield a stateful representation in the sense of \S\ref{subsec:critique-representation}: the resulting ``state'' is not a sufficient statistic for the dynamics being modeled, so prediction degrades regardless of predictor capacity.
A closed-loop generative architecture provides a built-in mechanism to detect this: if the decoder cannot reconstruct the next observation from the predicted state, the encoder was too lossy. The decoder thus serves not merely as a generation module but as a diagnostic tool that continuously pressures the representation to retain all dynamically relevant information. 
Without it, supervision occurs solely in the latent space, trading challenges of pixel-level variability for the risk of indefinability: the predicted latents are not directly grounded in observable data, which makes it difficult to diagnose whether the model is learning meaningful dynamics or collapsing to trivial solutions, an issue we discuss formally in \S\ref{subsec:objective}.



More broadly, the architectural divide between open-loop and closed-loop approaches reflects a deeper split in the representation learning community between \textit{semantic} representations (e.g., CLIP~\cite{radford2021learning} and DINO series~\cite{caron2021emerging,simeoni2025dinov3}), which are optimized to capture invariances across observations and discard the rest, and \textit{generative} representations (e.g., VAE~\cite{kingma2013auto}, MAE~\cite{he2022masked}, and diffusion models~\cite{ho2020denoising,rombach2022high}), which preserve enough variation to reconstruct the full observation space. JEPA's encoder is semantic by design due to its energy-based loss functions with heuristics-based regularization, which is powerful when the relevant semantics are known in advance, but limiting when the world model must generalize to tasks whose relevant features were not anticipated. 

On the other hand, generative representations subsume semantic ones: a sufficiently powerful generative model can develop rich semantic structure as a byproduct of learning to reconstruct observations, as demonstrated by LLMs, which acquire deep semantic understanding through next-token prediction. 
For a general-purpose world model, therefore, this insight supports adopting a generative architecture: the completeness of generative representations ensures that all dynamically relevant information is retained, while semantic structure can be organized by a dedicated reasoning layer (e.g., an LLM) that has already internalized language-based semantic structure through large-scale pretraining. Information discarded by a semantic encoder cannot be recovered, but information preserved by a generative encoder can always be further abstracted.
This complementarity matters especially because what a semantic encoder treats as ``important'' is determined by its training distribution: phenomena that are rare in the training data (e.g., car crashes in autonomous driving datasets) will be abstracted away, even when they are precisely the scenarios where the world model's predictions carry the highest stakes. 

Finally, a practical argument for abstraction-first architectures has been efficiency due to obviating the need of a separate decoder $g$, but recent advances are narrowing this gap. Systems such as FastVideo~\cite{zhang2025fast,zhang2026faster}, which combines trainable sparse attention with an LLM-based reasoning backbone, can generate 30 seconds of 1080P video in approximately 3 seconds, while architectures such as Causal Swin-DPM~\cite{xiang2025pan} use chunk-wise causal attention to dampen error propagation and signal variability over long horizons. These developments suggest that the marginal efficiency gain of dropping the decoder does not justify the fundamental information loss.


\subsubsection{Generative Latent Prediction as a Synthesis}

Therefore, rather than abandoning generative modeling to avoid signal variability, an alternative and well-established strategy is to adopt \textbf{hierarchical abstraction} through what we call a \underline{Generative Latent Prediction (GLP)} architecture (Figure~\ref{fig:critique-arch}, right).
Concretely, GLP consists of an \textbf{Encoder} $h$ and \textbf{Decoder} $g$ that together form a generative bottleneck grounding predictions in observable data, plus a \textbf{Latent Reasoning Backbone} $f$ composed of an enhanced LLM for discrete concept-based reasoning and a diffusion-based next-embedding predictor for continuous perceptual dynamics. 
Instead of modeling the full world at a single level of detail, GLP decomposes the problem across multiple layers of latent prediction, each specialized for different representational granularities, whether it be continuous perceptual features or discrete conceptual tokens. This allows each layer to operate at an appropriate level of abstraction while remaining generative and predictive.
For instance: 
\begin{itemize}
    \item At the lowest level, \textbf{next-embedding predictors} (e.g., latent diffusion models) can handle stochasticity and fine-grained variation in raw, continuous perceptual data (e.g., pixels, audio, proprioception). These models incorporate generative mechanisms (e.g., encoder-decoder architecture) that directly ground predictions in observable data, which leads to stronger supervision as we show in \S\ref{subsec:objective}.
    \item At the intermediate level, a \textbf{next-token predictor} (e.g., autoregressive Transformer decoder) can reason over discrete modality tokens derived via VQ-VAE-style encoders, capturing the symbolic and compositional structure.
    \item At the highest level, a \textbf{large language model (LLM)} operating in a ``thought space" composed of language tokens can support long-horizon planning, mental simulation, and counterfactual reasoning. Together with the intermediate level, these two levels of discrete reasoning can be jointly implemented through an \underline{enhanced LLM architecture} performing next-token prediction. 
\end{itemize}
The GLP paradigm not only supports structured, abstract reasoning through next-latent prediction, but also preserves the capacity for detailed reconstruction of the input world, enabling generative supervision and external use. This not only mitigates the compounding of prediction errors by isolating low-level variability within the bottom encoder-decoder layer, but also enables more expressive reasoning and generalization at higher layers of abstraction.
Importantly, it allows the model to flexibly mix continuous embeddings for perceptual nuance with discrete tokens for abstract structure, which aligns with our discussion of representation in \S\ref{subsec:critique-representation}.

Crucially, GLP does not reject semantic abstraction, but harnesses the fact that generative pretraining naturally gives rise to semantic structure. The LLM backbone, itself a product of large-scale generative pretraining, provides structured abstractions ready-made, while the generative decoder ensures \textit{completeness} by retaining information the semantic layer may not have known it needed. The result is a system where semantic abstraction is made \textit{sufficient} by a dedicated reasoning backbone and kept \textit{rich} by the generative decoder. In this sense, GLP is an inclusive paradigm: it naturally accommodates the growing family of efficient generative techniques, including FastVideo~\cite{zhang2025fast,zhang2026faster} and Causal Swin-DPM~\cite{xiang2025pan}, which make closed-loop validation practical at scale, while leveraging the powerful abstract reasoning capabilities of LLMs.
As we further elaborate in \S\ref{subsec:objective}, this encoder-world-model-decoder design leads to stronger supervision and more stable training dynamics than open-loop approaches, a claim we support formally in Propositions~\ref{prop:collapse-of-latent-loss}--\ref{prop:non-collapse-of-gen-loss} and Theorem~\ref{thm:latent-lower-bound-generative}.


\subsection{Objective: Learning in Data Space, or Latent Space?}
\label{subsec:objective}




What learning objective should guide world model training? Recent models (e.g., 1XWM~\cite{ho20251x} and LingBot-World~\cite{team2026advancing}) typically supervise the prediction based on reconstruction error against the observation data. An alternative is latent reconstruction (e.g., JEPA series~\cite{bardes2024revisiting,assran2025vjepa2selfsupervisedvideo}), which supervises transitions in the learned embedding space rather than reconstructing raw observations, motivated by tractability and the desire to avoid modeling irrelevant details.

A key argument for adopting latent supervision is that modeling a probabilistic distribution over raw observations (e.g., pixels in a video) for reconstruction may be unnecessary or counterproductive, and that learning in latent space can be more effective. This perspective has given rise to energy-based \textbf{latent reconstruction objectives}, which bypass the decoder and directly supervise transitions between encoded states. Formally, given encoder $h$ and world model $f$, the latent reconstruction loss is defined as:
\begin{equation}
    \mathcal{L}_{\text{latent}}(h, f) = \mathbb{E}_{(o, a, o') \sim \mathcal{D}} \left[ \left\lVert f(h(o), a) - h(o') \right\rVert \right],
\end{equation}
where the model predicts the next latent state $\hat{s}'$ and compares it to the encoded form of the next observation, without reconstruction $o'$ itself.

Despite its apparent simplicity, this objective is prone to collapse, as we show in Proposition~\ref{prop:collapse-of-latent-loss}: the model can trivially minimize the loss by mapping all observations to a constant vector and learning an invariant transition. 
Such collapse is the limiting case of statelessness (\S\ref{subsec:critique-representation}): the learned representation retains no identifiable information about the world state at all.
To counteract this tendency, latent reconstruction objectives often require complex regularizers (e.g., maximizing the information $I(\hat{s})$ of latent states). These regularizers, however, are often hard to tune and difficult to understand, which can make training brittle and limits scalability.
By contrast, the \textbf{generative reconstruction loss} grounds the learning objective in observable data by introducing a decoder $g$, and supervising the predicted next observation directly as below:
\begin{equation}
    \mathcal{L}_{\text{gen}}(h, f, g) = \mathbb{E}_{(o, a, o') \sim \mathcal{D}} \left[ \left\lVert g \circ f(h(o), a) - o' \right\rVert \right].
    \label{eq:gen-loss}
\end{equation}
Indeed, the generative loss $\mathcal{L}_{\text{gen}}$ anchors the learned representation to the structure of the sensory world, and thus avoids the collapse suffered by the latent loss $\mathcal{L}_{\text{latent}}$, as we show in Proposition~\ref{prop:non-collapse-of-gen-loss}.

\begin{figure}[t]
    \centering
    \includegraphics[width=1\linewidth,page=3]{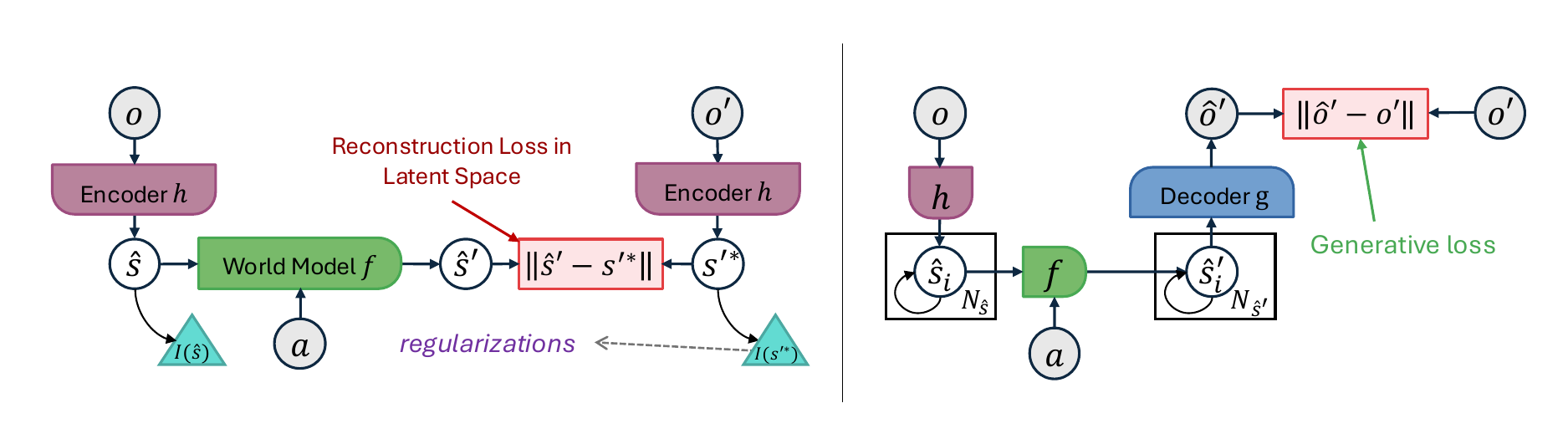}
    \caption{Comparison of latent-space reconstruction objectives (left) and generative data reconstruction objectives (right).}
    \label{fig:critique-obj}
\end{figure}

\begin{proposition}[Collapse of Latent Reconstruction Loss]
    \label{prop:collapse-of-latent-loss}
    Given $\mathcal{O},\ \mathcal{S},\ \mathcal{A} \subseteq \mathbb{R}^d$ and functions $h: \mathcal{O} \rightarrow \mathcal{S}$, $f: \mathcal{S} \times \mathcal{A} \rightarrow \mathcal{S}$, and latent reconstruction loss: $$\mathcal{L}_{\text{latent}}(h, f) = \mathbb{E}_{(o, a, o') \sim \mathcal{D}} \left[ \left\lVert f(h(o), a) - h(o') \right\rVert \right],$$
    There exists $(h^*, f^*)$ and $c \in \mathcal{S}$, such that $h^*(o) = c$ for all $o \in \mathcal{O}$ and $f^*(c, a) = c$ for all $a \in \mathcal{A}$, such that: $$\mathcal{L}_{\text{latent}}(h^*, f^*) = \min_{h, f} \mathcal{L}_{\text{latent}}(h, f)$$
\end{proposition}
\begin{explanation}
    If you have an encoder and a world model, and you train it using the latent reconstruction loss, there is a cheat configuration for the model to minimize the loss while learning nothing about the true dynamics.
\end{explanation}
\begin{proof}[Proof Sketch]
    If we construct such a degenerate solution $(h^*, f^*)$, this solution satisfies $\mathcal{L}_{\text{latent}}(h^*, f^*) = 0$, which is a global minimum as $\mathcal{L}_{\text{latent}}(h, f) \geq 0$ for all $(h, f)$.
\end{proof}
\begin{proposition}[Non-Collapse of Generative Loss]
    \label{prop:non-collapse-of-gen-loss}
    Given functions $h: \mathcal{O} \rightarrow \mathcal{S}$, $f: \mathcal{S} \times \mathcal{A} \rightarrow \mathcal{S}$, $g: \mathcal{S} \to \mathcal{O}$, and generative loss: $$\mathcal{L}_{\text{gen}}(h, f, g) = \mathbb{E}_{(o, a, o') \sim \mathcal{D}} \left[ \left\lVert g \circ f(h(o), a) - o' \right\rVert \right],$$
    Assuming $\exists (o_1, a_1, o_2),\ (o_3, a_3, o_4) \in \mathcal{D}$ such that $o_2 \neq o_4$, then given $(h', f')$, fixed $g'$ and $c \in \mathcal{S}$ such that $h'(o) = c \ \forall o \in \mathcal{O}$ and $f'(c, a) = c \ \forall a \in \mathcal{A}$, there exists $(\tilde{h}, \tilde{f})$ such that: $$\mathcal{L}_{\text{gen}}(\tilde{h}, \tilde{f}, g') < \mathcal{L}_{\text{gen}}(h', f', g')$$
\end{proposition}
\begin{explanation}
    If you add a decoder to the model and train it with the generative loss, assuming the data contains different next-observation targets, there will always be another set of encoder and world model that gets lower loss than the previous cheat configuration.
\end{explanation}
\begin{proof}[Proof Sketch]
    Given degenerate solution $(h', f')$ and fixed $g'$, construct $(\tilde{h}, \tilde{f})$ to be equal to $(h', f')$ at every point except $(o_1, a_1, o_2)$ and $(o_3, a_3, o_4)$ where $o_2 \neq o_4$ and the constant-valued $(h', f')$ will get non-zero loss. Instead, set $(\tilde{h}, \tilde{f})$ to perfectly fit these two targets, so they will get zero loss. Thus we have:
    $$\mathcal{L}_{\text{gen}}(\tilde{h}, \tilde{f}, g') < \mathcal{L}_{\text{gen}}(h', f', g')$$
\end{proof}
Details of the proofs of both propositions are available in Appendices~\ref{appendix:proof-collapse-of-latent-loss} and \ref{appendix:proof-non-collapse-of-gen-loss}, respectively.

Beyond the issue of collapsing, a more fundamental structural limitation of the latent reconstruction objective is that it essentially acts as a loose surrogate for observation-level consistency, as we show in Theorem~\ref{thm:latent-lower-bound-generative}. 
This means that minimizing  $\mathcal{L}_{\text{latent}}$ does not, in general, guarantee consistency with what the agent would observe in the world, which can lead to misaligned or brittle representations. In general-purpose settings, we argue that anchoring to the next observation $o'$ via generative loss provides a more stable and mechanistically interpretable training signal.

\begin{figure}[t]
    \centering
    \includegraphics[width=0.8\linewidth,page=2]{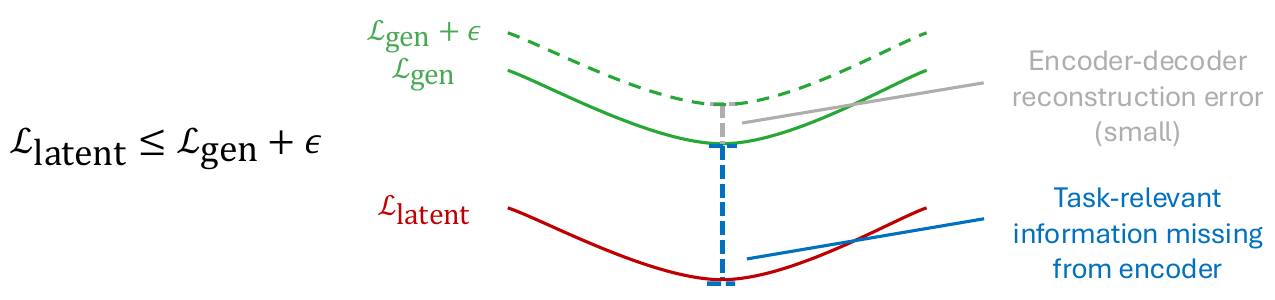}
    \caption{As Thm.~\ref{thm:latent-lower-bound-generative} shows, the energy-based \textit{latent} reconstruction loss ($\mathcal{L}_{\text{latent}}$) is upper bounded by the generative \textit{data} reconstruction loss ($\mathcal{L}_{\text{gen}}$) plus a small encoder-decoder reconstruction error ($\epsilon$). $\epsilon$ is small in practice, meaning $\mathcal{L}_{\text{latent}} \leq \mathcal{L}_{\text{gen}}$ usually holds. Minimizing $\mathcal{L}_{\text{latent}}$, therefore, does not guarantee consistency with observed data, which is required for minimizing $\mathcal{L}_{\text{gen}}$. 
    }
    \label{fig:latent-gen-proof}
\end{figure}

\begin{theorem}[Latent reconstruction is an upper-bounded surrogate for generative reconstruction]
\label{thm:latent-lower-bound-generative}
    Given sufficiently powerful encoder $h: \mathcal{O} \to \mathcal{S}$ and decoder $g: \mathcal{S} \to \mathcal{O}$, such that for all latent states $\hat{s} \in \mathcal{S}$, the roundtrip reconstruction error satisfies $\left\lVert h \circ g(\hat{s}) - \hat{s} \right\rVert \leq \epsilon$ for some small $\epsilon > 0$. For world model $f: \mathcal{S} \times \mathcal{A} \to \mathcal{S}$ and transition data $(o, a, o') \sim \mathcal{D}$, define the latent-space loss $\mathcal{L}_{\text{latent}}$ and the generative loss $\mathcal{L}_{\text{gen}}$ as below: 
    \[
        \mathcal{L}_{\text{latent}} = \left\lVert f(h(o), a) - h(o') \right\rVert,
        \qquad
        \mathcal{L}_{\text{gen}} = \left\lVert g \circ f(h(o), a) - o' \right\rVert.
    \]
    Assume that encoder $h$, decoder $g$, and world model $f$ induce the following conditional distributions:
    \[
        \hat{s} \mid o \sim \mathcal{N}(h(o), I),
        \quad
        \hat{o} \mid \hat{s} \sim \mathcal{N}(g(\hat{s}), I),
        \quad
        \text{and }
        \hat{s}' \mid \hat{s}, a \sim       \mathcal{N}(f(\hat{s}, a), I).
    \]
    Then, the following inequality holds:
    \[
        \mathcal{L}_{\text{latent}} \leq \mathcal{L}_{\text{gen}} + \epsilon,
    \]
    with $\mathcal{L}_{\text{latent}} = \mathcal{L}_{\text{gen}}$ 
    when $h \circ g(\hat{s}) = \hat{s}$ for all $\hat{s} \in \mathcal{S}$ and $\supp(\mathcal{D}) \subseteq \image(g)$.
\end{theorem}
\begin{explanation}
    If your encoder and decoder approximately undo each other, then the JEPA latent reconstruction loss is upper-bounded by the generative loss plus a small reconstruction error. These losses are only equal when the encoder-decoder pair perfectly invert each other, which is unrealistic in practice.  
    As such, \underline{minimizing the latent loss does not guarantee consistency with observed data}, which is required for minimizing the generative loss (Figure~\ref{fig:latent-gen-proof}).
\end{explanation}
\begin{proof}[Proof Sketch]
Observe that the two losses $\mathcal{L}_{\text{latent}}$ and $\mathcal{L}_{\text{gen}}$ are scaled KL divergences of the Gaussian prediction distributions in the latent and observation spaces, respectively. Apply the encoder $h$ to both the observation prediction and true observation distributions in $\mathcal{L}_{\text{gen}}$, and the \textbf{data processing inequality} states that the augmented KL divergence is upper-bounded by the original $\mathcal{L}_{\text{gen}}$. After that, apply the triangle inequality to show that $\mathcal{L}_{\text{latent}}$ is upper-bounded by the sum of this augmented KL (upper-bounded by $\mathcal{L}_{\text{gen}}$) and the roundtrip reconstruction error (upper-bounded by $\epsilon$), thus completing the proof. (We include the detailed proof in Appendix~\ref{appendix:latent-lower-bound-generative}.)
\end{proof}

In practice, $\epsilon$ is small (as is typical for strong modern autoencoders), so $\mathcal{L}_{\text{latent}} \leq \mathcal{L}_{\text{gen}}$ usually holds, meaning that the former can miss semantically important mistakes that the latter will penalize. Additionally, the use of less-understood regularizers in conjunction with $\mathcal{L}_{\text{latent}}$ as the objective makes its outcome even more difficult to assess without  necessary boundary conditions imposed by observation data. 

More broadly, all representation necessarily involves compression; the question is not whether to compress, but how to avoid premature loss of task-relevant information. The emphasis on learning abstract, predictable structure (e.g., JEPA) is well-founded, as a model should indeed focus on what is learnable and avoid squandering capacity on irrelevant noise~\cite{lecun_xing_world_models_2026}. 
Our perspective extends this insight by arguing that generative objectives provide a natural safeguard: by requiring the model to reconstruct observations, the learning process itself determines what information is worth retaining, rather than relying on \textit{a priori} design decisions about which details matter.

In conclusion, our argument is not that world models must operate in pixel space, but that they should learn from it. 
Framing the distinction as \textit{next-representation prediction versus next-observation prediction} creates a false dichotomy that can lead to theoretical ambiguities and practical instability.
The purpose of predicting the next observation is to ensure that the predicted latent representations are meaningfully grounded in the real world, whether conceptually or physically. Conversely, reliable prediction in latent space depends on continual validation through observable data.
Mathematically, any latent representation of real-world signal intrinsically suffers from issues of identifiability and stability. As such, alignment and calibration with real data are essential to ensure the representations remain meaningful and robust. 
Generative reconstruction objectives tether the learned representations to the observable world, providing richer and more stable learning signal that supports meaningful distinctions, general usability, and human interpretability. These properties are critical for downstream usage, whether it be planning trajectories or training agents through reinforcement learning, which we discuss more in \S\ref{subsec:usage} below.

\subsection{Usage: MPC or RL?}
\label{subsec:usage}



\begin{figure}[t]
    \centering
    \includegraphics[width=1\linewidth,page=4]{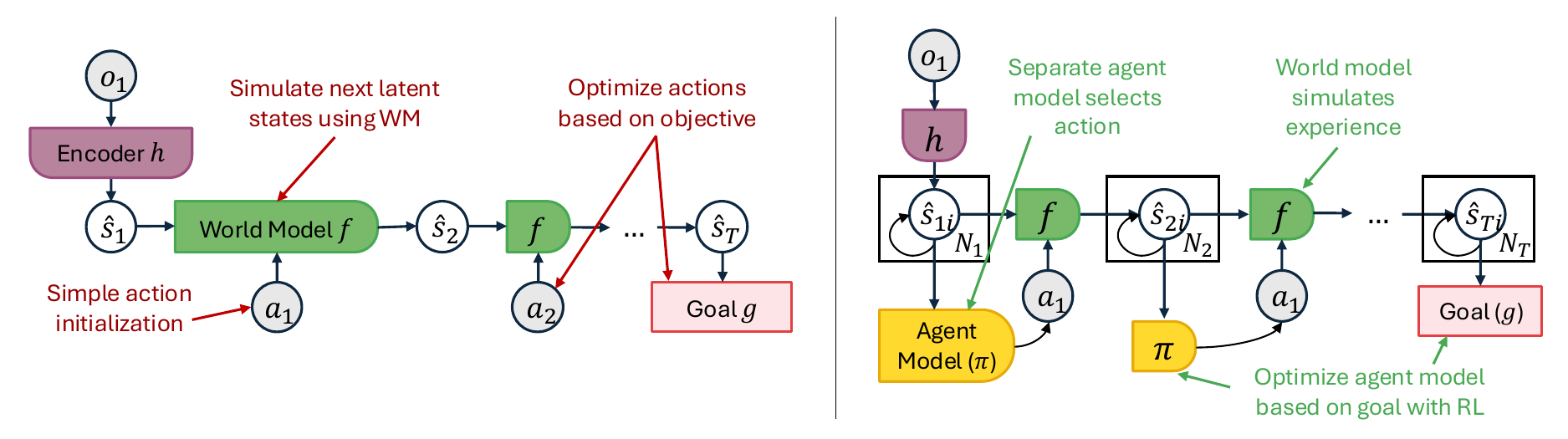}
    \caption{Comparison of model-predictive control (left) and reinforcement learning from world-model-simulated experience (right) as approaches to using the world model for decision-making.}
    \label{fig:critique-usage}
\end{figure}

How should a trained world model be used for decision-making? 
There has also been debate over whether model-predictive control (MPC) is favored over reinforcement learning (RL) in using the WM for reasoning, due to sample efficiency and safety advantages~\cite{finn2017deepvisualforesightplanning}.
Here we describe a typical MPC setup (Figure~\ref{fig:critique-usage}, left) which is often adopted by recent work~\cite{sobal2025learning,assran2025vjepa2selfsupervisedvideo}: at timestep $t$ during inference, the agent infers its current latent state $\hat s_t = h(o_t)$, 
proposes an initial sequence of actions $(a_t, \dots, a_{T-1})$ until some decision horizon $T$, and uses the world model to predict the corresponding next-state sequence $(\hat{s}_{t+1}, \dots, \hat{s}_{T})$. These simulated states can then be evaluated using a cost function $C(g, \hat{s})$ for goal $g$ (e.g., L2 distance between $\hat{s}$ and encoded goal $\hat{s}_g = h(g)$), based on which the agent may propose the next action with lower cost. Decision-making thus amounts to finding the action sequence that minimizes the cost function, formalized as below:
\[
    (a_t^*, \dots, a_{T-1}^*) = \argmin_{a_t, \dots, a_{T-1}} \sum_{k=t}^{T-1} C(g, f(\hat{s}_k, a_k)).
\]
In practice, the (continuous) action optimization is often performed using traditional numerical algorithms (e.g., MPPI~\cite{williams2015model} and CEM~\cite{RUBINSTEIN199789}) involving decision horizons of 1-20 steps and 100s upon 1000s of action samples, and the agent executes the first action in the final action sequence $a_t^*$ before replanning in the next step $t+1$.
The appeal of MPC lies in learning from offline trajectories  $(o_1, a_1, \dots, o_T) \sim \mathcal{D}$ without potentially unsafe exploration in the real world, as well as potential for higher-quality decision-making from world-model-based simulation.

However, MPC can suffer from practical limitations.
Simulation of latent trajectories using the world model, for instance, must be performed repeatedly at every timestep during inference, leading to high computational overhead and making it difficult to respond effectively in fast-changing environments. 
Beyond computational efficiency, MPC typically plans only a few steps ahead (e.g., up to 10-20 steps) in terms of searching horizon. This limits the extent of its foresight, as long planning horizons (e.g., 100s of steps) can be difficult due to the exploding number of trajectories and world-model errors. 
As the horizon increases, MPC also becomes more difficult to implement and optimize, since the proposal distribution must sample entire action sequences at once over the full planning horizon. This is why MPC often relies on relatively simple proposal distributions, such as uniform random sampling or multivariate Gaussians. Indeed, MPC so far has shown promise primarily in simplified settings (e.g., Go) where environment dynamics are simple and slower decision-making is rewarded, but struggles to extend to real-world tasks (e.g., customer service), which typically involve complex dynamics and require a mixture of short- and long-term decision-making.

On the other hand, RL is a general, flexible, and scalable approach to training agents without restrictions on the decision-making method or search horizon. In particular, one can replace the true universe with a world model $f$ for exploration and learning (as discussed in \S\ref{subsec:world-model}). Below we describe an RL setup (Figure~\ref{fig:critique-usage}, right) where the agent interacts with the world model instead of the environment~\cite{ha2018world}: in each time step $t$ with world state representation $\hat{s}_t$ (which may be encoded from some observation data $o_t$ or completely imagined from scratch), the agent $\pi$ takes action $a_t \sim p_\pi(a_t \mid \hat{s}_t)$ and the world model $f$ simulates the next state $\hat{s}_{t+1} \sim p_f(\hat{s}_{t+1} \mid \hat{s}_t, a_t)$. This may repeat until some rollout horizon $T$ or in a never-ending manner. Computing the reward at each step $r(g, \hat{s}_t)$ based on goal $g$, the optimal agent $\pi^*_f$ may thus learn by maximizing the expected discounted cumulative reward (with well-formed discount schedule $\{ \gamma_k \}_{k=t}^\infty$ to ensure numerical stability) as defined below:
\begin{align}
    \label{eq:rl-world-model}
    \pi^*_f(\hat{s}_t) &= \argmax_{\pi} \mathbb{E}_{\pi, f} \left[ \sum_{k=t}^\infty \gamma_k r(g, \hat{s}_k) \mathrel{\bigg|} \hat{s}_t \right] \\
    &= \argmax_{\pi} \lim_{T \to \infty} \sum_{(a_t, \hat{s}_{t+1}, \dots, \hat{s}_T)}  \sum_{k=t}^T \gamma_k r(g, \hat{s}_k) \ \prod_{i=t}^T {\underbrace {p_\pi(a_i \mid \hat{s}_i)}_{\scriptsize \shortstack{select action}}} \ {\underbrace {p_f(\hat{s}_{i+1} \mid \hat{s}_i, a_i).}_{\scriptsize \shortstack{simulation with \\ world model}}} \nonumber
\end{align}

Operationally, as we show above, both MPC and RL can use world models, the former only for decision-making while the latter also for learning. We recognize the latter as part of a broader paradigm: \textbf{learning from experience}~\cite{hu2021toward}. In this framework, the agent model continually interacts with and learns from an infinite space of imagined worlds, simulated by a world model. The countless hypothetical trajectories can then be used to train the agent via RL, imitation learning, or other learning signals that make full use of all experiences. 
These updates can occur entirely offline, using batches of rollouts from the world model rather than interacting with the real environment.

Compared to MPC which is computationally expensive at decision-making time, RL with world model (as in Equation~\ref{eq:rl-world-model}) shifts part of the computational cost to the training phase. Instead of planning from scratch at each step, the world model is used offline to train a policy network that can later be reused for fast action selection at every state. 
Crucially, Both RL and MPC along with the world model may be included as components inside the agent model that must carry both deliberate planning and reactive actioning, while another fast policy can still learn to react swiftly when needed. Whereas recent work like o1, o3, and R1~\cite{guo2025deepseek} can be seen as special cases in math and coding where model-free policy-based methods enables fast-reacting behaviors, our view is to generalize this pattern: Agents should both reason with and learn from the worlds that they simulate, allowing flexible decision-making, continual improvement, and emergence of intelligence with experience.

In summary, as demonstrated above, unlike MPC, RL can learn a policy function that reflects long-term cumulative rewards, enabling more strategic reasoning over extended time horizons. This makes it applicable in practical settings like goal-conditioned robotic manipulation, multi-turn dialog systems, or autonomous driving.

\section{The PAN World Model}
\label{sec:pan-world-model}

Drawing from the examination of world model design dimensions above, we arrive at the following conclusion regarding design principles for a general-purpose WM capable of supporting simulative reasoning: 1) use data from all modalities of
experience; 2) employ a 
stateful,
mixed continuous and discrete representation; 3) adopt a hierarchical Generative Latent Prediction (GLP) modeling paradigm
with an extended-LLM backbone (for discrete concept-based reasoning), as well as a generative embedding predictive module (for continuous gradient-based reasoning), as the reasoning engines; 4) train over a generative loss grounded in observation data; and 5) apply world model
to simulate experiences for training agents using reinforcement learning. We conclude our examination with a brief preview of a new architecture, PAN -- a Physical,
Agentic, and Nested world model, based on the aforementioned design principles. Details and preliminary results of PAN will be presented in~\cite{xiang2025pan}.

\begin{figure}[t]
    \centering
    \includegraphics[width=1.0\linewidth,page=3]{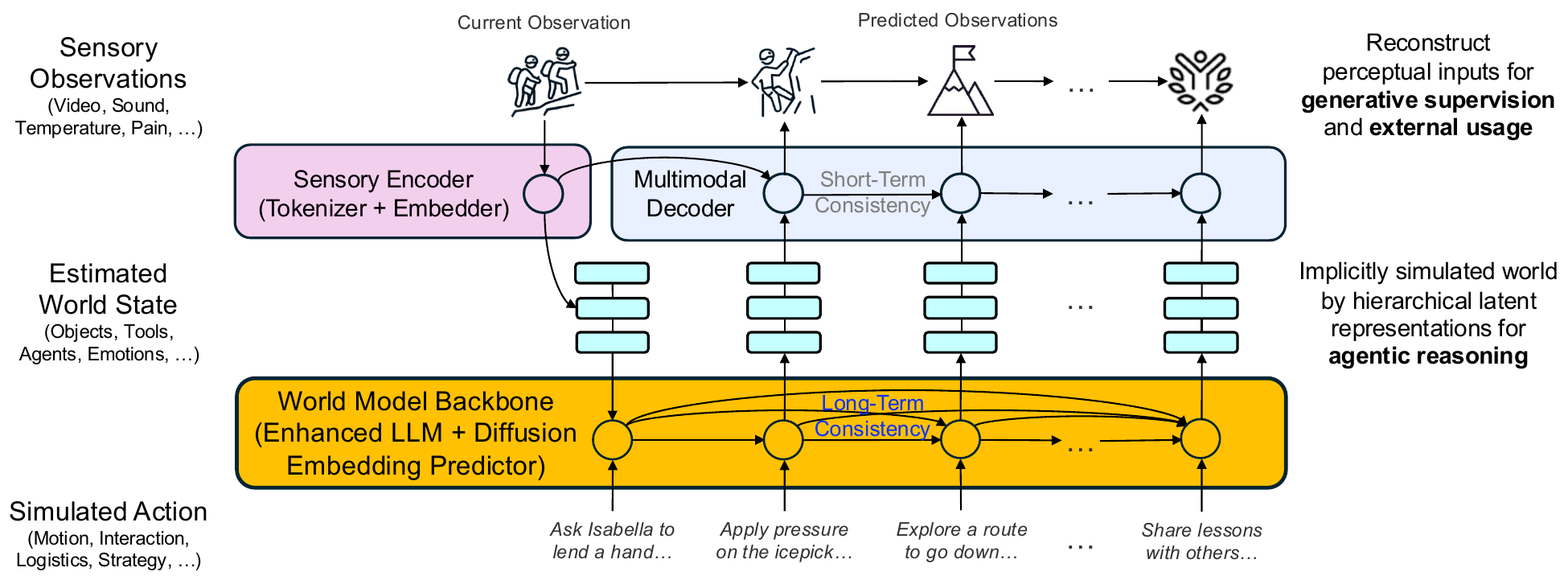}
    \caption{Illustration of the generative latent prediction (GLP) architecture of PAN-World, our proposed framework for general-purpose world modeling. At each time point, PAN-World estimates the world state $\hat{s}$ based on the current sensory observation $o$, predicts the future world states $\hat{s}'$ based on proposed actions $a$ for agentic reasoning, and reconstructs the future observations $\hat{o}'$ for generative supervision and external usage.}
    \label{fig:pan-arch}
\end{figure}

\subsection{A Motivating Usecase}

A truly versatile and generalizable WM must be grounded in tasks that reflect the full complexity of real-world reasoning demands. These may include variations in data modality (e.g., verbal, visual, sensory), spatio-temporal scope (from one second in a room to days in a whole country), action granularity (e.g., fine motor control, bodily movement, expressive gestures), and decision scale (from immediate actions, tactics, to long-term strategies).
While many existing WMs are demonstrated on simplified, toy tasks (e.g., manipulating kitchen tools) and simple scenarios (seconds to minutes of video in 3D worlds), such settings fall short of capturing the richness of real-world agentic experience. A WM designed around these tasks, therefore, is unlikely to scale to complexities required for real-world applications. For instance, a WM that only enables tool manipulation in a kitchen is inadequate for planning and executing an end-to-end dinner service in a restaurant.

In contrast, PAN is motivated by a more complex and realistic use case: a mountaineering expedition. In this setting, the WM must internalize multimodal sensory inputs and simulate future world states in service of a demanding, structured task. This task naturally decomposes into multiple interconnected sub-tasks at different levels: high-level decisions like gear selection, route and segment planning, navigation, weather assessment, pacing, etc., low-level actions like climbing, roping, and precise motor control in response to terrain and surface conditions, and social coordination with teammates through verbal and non-verbal communication, among others.

The mountaineer's sensory experience includes not only sights and sounds -- snowfields, cliffs, a partner calling out from ahead -- but also tactile and motion signals such as wind, cold, and muscular strain. The actionable world states that drive purposeful reasoning, such as terrain affordances, team dynamics, or latent risks, exist at multiple levels of abstraction beneath them. PAN thus begins by ingesting this continuous stream of multimodal signals: inputs from vision, sound, temperature, motion, potentially even pain, which may each be relevant for different tasks but together constitute a holistic reality.

\subsection{The GLP Architecture for PAN World Model}

Following a mixed representation and multi-scale reasoning principle, PAN processes multimodal sensory inputs $o$ using its \underline{Sensory Encoder ($h$)}, which maps inputs via both discrete and continuous pathways to capture complementary aspects of the world. 
On one hand (Figure~\ref{fig:pan-arch}), a tokenizer hierarchically maps raw signals into discrete tokens grounded in PAN's vocabulary, which spans multiple levels of abstraction. These tokens may consist of abstract tokens learned via a VQ-VAE-style approach~\cite{van2017neural} as well as concrete words drawn from natural language. 
The representation may include a flexible number of such tokens to compactly reflect deep layers of world information: \textit{Where am I? Who is with me? What tools do I have? What is my emotional state?} 
This token-based component of the world state is \textit{stateful} by construction: grounded in a persistent vocabulary, each token is identifiable and re-addressable, allowing the estimated state to be held in memory, queried, and updated across the long horizons of the expedition rather than recomputed transiently from each new sensory frame.
As discussed in \S\ref{subsec:critique-representation}, this form of representation can be sufficient to capture relevant information, even for continuous data like video. 
On the other hand (Figure~\ref{fig:critique-arch}, right), PAN may also encode low-level details into continuous latent embeddings to capture the full nuanced perceptual experience where necessary. 
Together, these tokens and embeddings form a layered estimate of the world state $\hat{s} = \{\hat{s}_{i}\}_{i=1}^{N}$ over which PAN performs simulation and purposeful reasoning.

Given a proposed action $a$ (e.g., ``buckle the carabiner to my harness''), PAN predicts the next world state $\hat{s}'$ (e.g., a conceptual state like ``I am safely anchored'', or a physical state like ``that rope is tightening'') using a \underline{World Model Backbone ($f$)} built on an enhanced LLM and a diffusion-based next-latent-embedding predictor (Figure~\ref{fig:pan-arch}). This design is a concrete instantiation of the GLP architecture introduced earlier in \S\ref{subsec:critique-architecture} (Figure~\ref{fig:critique-arch}, right).
The LLM-based backbone reasons over both natural language tokens and a learned conceptual vocabulary -- some explicit (e.g., a particular shape of icepicks), others implicit or emergent (e.g., feelings that arise when sharing hard-earned knowledge).
This supports broad generalization across domains~\cite{simura2025}.
During both training and inference, the model can also dynamically extend its vocabulary by introducing new tokens or merging existing ones to maximize the prediction quality. 

On the other hand, the diffusion-based embedding predictor is responsible for fast, low-level, and often subconscious reasoning that are critical for embodied responses yet difficult to express in language. 
This module simulates detailed perceptual experiences, such as whether a foothold is secured, or how the body might shift its weight during a climb~\cite{xiang2024pandora}. 
A \underline{Learned Switch} allows PAN to predict the next world state hierarchically ($\{\hat{s}'_{i}\}_{i=1}^{N'}$) by adaptively combining the LLM-based backbone, multiple vocabularies, and the diffusion-based embedding predictor, depending on task demands. 
These mechanisms enable PAN-WM to scale across spatio-temporal scopes and action granularities as is required for general usability -- from concrete physical scenarios like mountain climbing and social interaction, to abstract, far-reaching strategic consequences like nationwide policy changes. 

To supervise its predictions, and to allow the trained WM to interface with external agents (or human) who may use its outputs, PAN reconstructs the next observation $\hat{o}'$ using a \underline{Multimodal Decoder} \underline{($g$)} and compares it to the actual observation $o'$. Crucially, the decoder's outputs are not limited to videos, but includes a full sensory experience, which may include sound, temperature, motion, pain, other embodied signals, and/or even text.
As discussed in \S\ref{subsec:critique-architecture} and \S\ref{subsec:objective}, this generative supervision grounds the predicted world state $\hat{s}'$ in sensory reality, ensuring that the representation retains all possible information while allowing residual variability to be absorbed by the decoder $g$. This approach contrasts sharply with models trained on next-representation prediction (e.g., V-JEPA 2~\cite{assran2025vjepa2selfsupervisedvideo}), which supervise the world model purely in latent space. The latter objectives are, at best, loose surrogates of generative objectives and prone to representation collapse or unidentifiability, as they lack grounding in real sensory input.
Formally, PAN models the conditional distribution of the next observation $o'$ given the current observation $o$ and proposed action $a$ as below:
\begin{align*}
    &p_{\text{PAN}}(o' \mid o, a) \\
    = & \sum_{\hat{s}, \hat{s}'} {\underbrace {p_h(\hat{s} \mid o)}_{\scriptsize \shortstack{encoder}}} \ {\underbrace {p_f(\hat{s}' \mid \hat{s}, a)}_{\scriptsize \shortstack{world model}}} \ {\underbrace {p_g(o' \mid \hat{s}')}_{\scriptsize \shortstack{decoder}}} \\
    = & \sum_{\hat{s}, \hat{s}'} {\underbrace {\prod_{i=1}^N p_h(\hat{s}_i \mid \hat{s}_{<i}, o)}_{\scriptsize \shortstack{hierarchical world \\ state inference}}} \ {\underbrace {\prod_{j=1}^{N'} p_f(\hat{s}'_j \mid \hat{s}'_{<j}, \hat{s}, a)}_{\scriptsize \shortstack{switch-based next-state \\ prediction}} } \ p_g(o' \mid \hat{s}')
\end{align*}

Overall, with its hierarchical, multi-level, and mixed representation architecture, and an encoder-decoder pipeline that threads perception $o$, action $o$, belief $\hat{s}_i$, simulated belief $\hat{s}'_i$, and simulated worlds $o'$, PAN represents a general-purpose generative model for simulating actionable real-world possibilities for an agent to perform purposeful reasoning, as we will briefly allude to in $\S5.4$.
PAN does not sidestep the variability in raw perceptual input, but instead modularizes and organizes it. This enables richer internal simulation of every layer of experience for more powerful agent reasoning and planning.

\subsection{Training the PAN World Model}

It should be obvious from the mountaineering example that simply watching videos is not enough to learn all the reasoning capability needed to accomplish the final goal, which can take days of time and hundreds upon thousands of actions and steps from the onset, and is built on rich background knowledge about geography, climate, equipments, sports, and even history. The training of PAN-WM shall use a divide-and-conquer approach that begins with pretraining each of its modules independently through self-supervision (e.g., LLM for text data, and diffusion model for video data). 
These modality- and level-specific modules are then aligned or integrated during the post-training phase using multimodal data, cascaded embedding, and gradient propagation. 
Modules operating over continuous embeddings can be trained using standard gradient-based optimization techniques. In contrast, components using discrete tokens may benefit from gradient-free methods similar to reinforcement learning~\cite{guo2025deepseek}.
As proven in \S\ref{subsec:objective}, generative, data-reconstruction-based objectives are grounded in observed data and provide a stable and reliable learning signal for the entire system.

A key strength of the PAN architecture is its data efficiency, because of its use of a multi-scale and stratified view of the world. In the mountaineering task, when reasoning about navigation and path-finding, the world states do not need to include snow or rock surface details at pixel level, whereas when deciding where to lay hands or feet during climbing, the world states can ignore geographical contexts. Therefore it is not necessary for WMs simulating highly complex possibilities to be contingent on data that capture all such complexity all at once (e.g., videos that visually cover mountaineering at all levels), but to take advantage of data of different kinds offering information at different levels (e.g., travel book for trail guide and map reading, indoor video for rock climbing and gear usage).  
After all, it is unrealistic to expect a large corpus of videos that comprehensively cover all aspects of alpine climbing. 
Many general capabilities (e.g., social reasoning, travel planning, cold weather survival) can be learned from abundant language data.
Only directly embodied skills (e.g., foot placement, rock climbing technique) require physical data like videos or proprioception, which can be obtained in controlled or simulated environments.

Indeed, PAN's pretrain-then-align/integrate strategy enables sensory information (e.g., from a video diffusion model) to be grounded within higher-level, richer contexts through LLMs, thereby facilitating cross-modal generalization.
At the same time, abstract knowledge embedded in LLMs can be anchored to concrete, embodied experiences, increasing the precision and realism of the system's reasoning~\cite{xiang2023language}.
The result is a WM that, like humans, derives commonsense understanding from a diverse set of experiences. Consequently, it does not require exhaustive training data for each specific task (e.g., mountaineering or autonomous driving), but can instead draw on conceptual knowledge acquired from many domains.
We believe this kind of general-purpose WM is well-suited to simulating experience for agent decision-making and/or training, as elaborated below.

\subsection{Towards Agentic Reasoning with PAN}
\label{sec:agentic-reasoning}

\begin{figure}[t]
    \centering
    \includegraphics[width=1.0\linewidth,page=3]{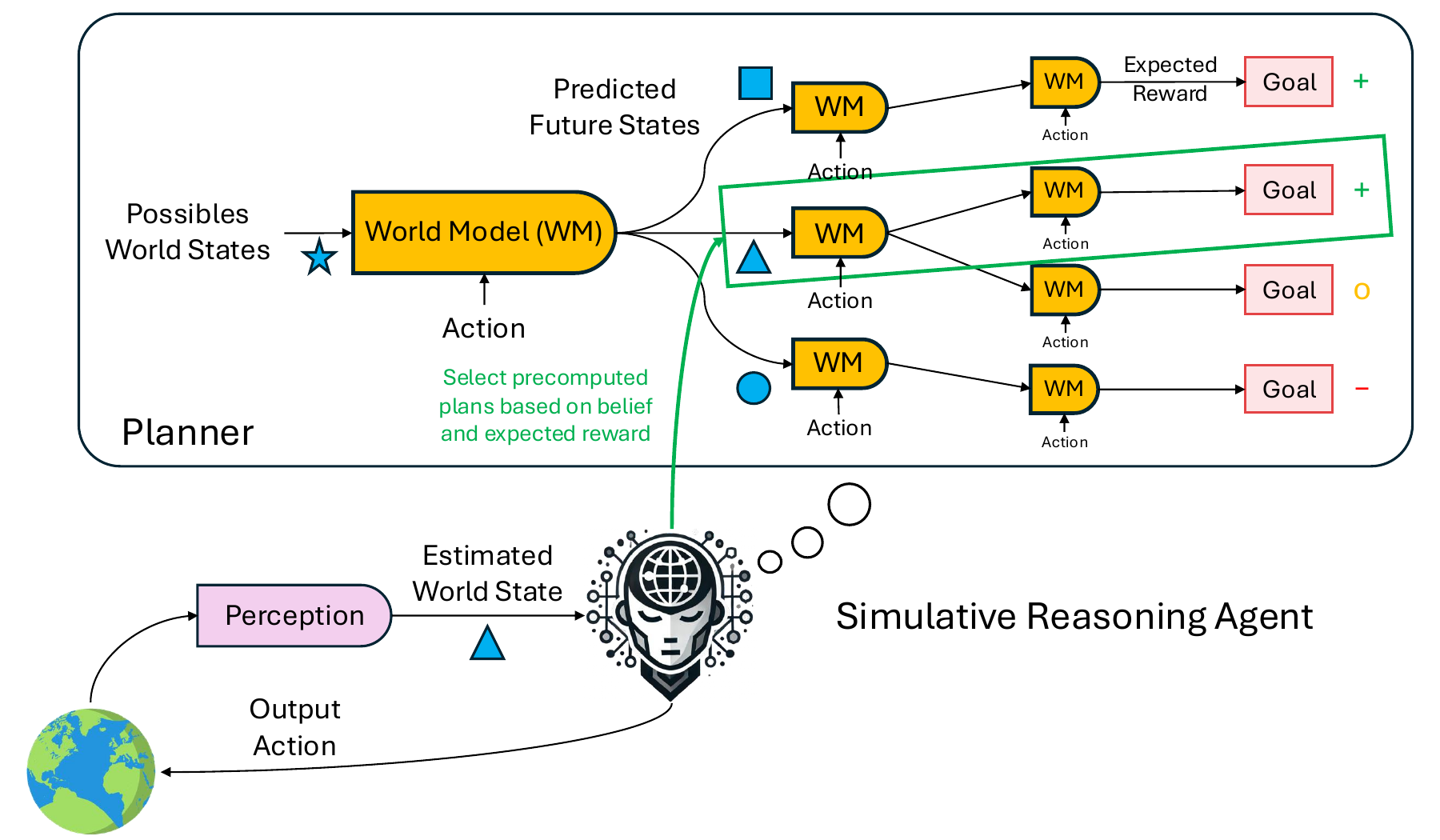}
    \caption{Illustration of our proposed simulative reasoning agent powered by the PAN World Model. Unlike traditional RL agents that rely on reactive policies, or model-precitive control (MPC) agents that expensively simulates futures at decision time, this agent leverages a cache of precomputed simulations generated by the PAN WM. During decision-making, the agent selects actions based on its current beliefs and expected outcomes, enabling a more efficient, flexible, and intentional form of planning, which, as we argue, is closer to the flexibility of human reasoning. }
    \label{fig:new-agent-model}
\end{figure}

Recall in $\S2$ we outlined an agent architecture for simulative reasoning using the world model. PAN 
fits naturally into this paradigm, functioning not merely as a video generator, but as a rich internal sandbox for simulation, experimentation, and foresight.

As illustrated in Figure~\ref{fig:new-agent-model}, a PAN-Agent, prompted by a goal and in reception of continuous stream of perception from the real world, is expected to come up with actions, plans (sequence of actions), or strategies (plans in light of counterfactual situations), which would involve using the PAN-WM to precompute and cache a diverse set of possible world states, plausible actions within those states, and their simulated outcomes~\cite{reasoneragent2025}.
At decision time, rather than only relying on performing expensive real-time simulations, the agent may consult this cache and select actions based on current beliefs and expected rewards.
This decoupling of simulation from action selection allows the agent to reason more deliberately, adaptively, and selectively, avoiding the rigidity of purely reactive policy in end-to-end RL and the computational burden of constant forward rollout in MPC. The result is an agent that more closely mirrors human-like cognition -- planning ahead, navigating uncertainty, and acting with both flexibility and foresight.
We believe this represents a promising step towards agents with richer forms of agency -- one capable not only of simulative reasoning, but also of choosing among imagined futures with intent. Such agents may ultimately approach the adaptability, resilience, and autonomy characteristic of human intelligence. 

\section{Conclusion}
We have examined the foundations, debates, and practical challenges in the pursuit of general-purpose world modeling. 
Our intent in writing this critique is to inspire further discussions and deeper reflections on the following fundamental questions: {\it what is indeed a world model?}, {\it what is a world model for}, and {\it how to build a world model of practical and general utility?} 

We argue that a WM is not about video or virtual reality generation, but is about simulating all possibilities in real world; and such outcome is not for visual pleasure, but for simulative reasoning; and current paradigms and efforts towards this end remain primitive. 

It is our wish that, by offering a constructive and analytical examination of the key dimensions of world model design, surveying the diverse perspectives in the field, and presenting our alternative proposal -- 
the Generative Latent Prediction (GLP) architecture,
we can spark further advancements in both theory and implementation of stronger world models.

The PAN world model we previewed is a concrete instantiation of GLP, designed for simulating all possible worlds and enabling agent reasoning and planning. By combining multimodal data, 
stateful,
hierarchical representations, multi-level generative modeling, and observation-grounded objectives, PAN supports long-horizon simulation and flexible decision-making across tasks in the physical and cyber domains.

It has been recognized that world models should serve as core components of agentic planning and decision-making, as echoed in recent work~\cite{li2026functional_taxonomy_world_models,nvidia2026cosmos3}. We, however, argue that how to \textit{act} on world-model simulations is the function of a separate agent model, as previewed in \S\ref{sec:agentic-reasoning}. Moreover, the agent's relationship to the world model goes well beyond planning and action selection, and we believe that these capabilities should arise as properties internal to the agent itself, rather than externally prescribed procedures. We address these questions in a forthcoming companion manuscript~\cite{xing2026cam}.

Looking ahead, the PAN framework opens several promising directions: scaling from single-agent to multi-agent simulations (e.g., collective behaviors of a business, a society, consequences to public health), extending across time scales (e.g., from milliseconds to millennia), improving simulation fidelity across modalities, and enabling agent learning directly through imagined experience. 
As world models increasingly serve as the substrate for reasoning, imagination, and action, we believe that frameworks like PAN, with its experience grounding, multi-layer abstraction, and empirical scalability, offers a compelling foundation for the development of robust, general-purpose AI.


\clearpage

\bibliographystyle{plain} 
\bibliography{refs}

\appendix 

\section{Proof for Theorem~\ref{thm:completeness-of-language}}
\label{appendix:proof-completeness-of-language}
\begin{proof}
We will prove the contrapositive that $f_\epsilon(\mathbf{x}) = f_\epsilon(\mathbf{x}') \Rightarrow \| \mathbf{x} - \mathbf{x}' \| \leq \epsilon$.

Based on the conditions, we have it that $\| \mathbf{x} - \mathbf{x}' \| = \sum_{t=1, d=1}^{T, D} (x_{td} - x'_{td})^2 \leq \epsilon = \Tilde{\epsilon}^2$, which is satisfied when $|x_{td} - x'_{td}| \leq \frac{\tilde{\epsilon}}{\sqrt{TD}}$.

Because $\| x_t \| \leq K \vcentcolon= \frac{\Tilde{K}^2}{4}$, we have that $| x_{td} | \leq \frac{\Tilde{K}}{2} \ \forall d \in [1, D]$. Therefore, to satisfy the condition that $\| \mathbf{x} - \mathbf{x}' \| \leq \epsilon$, we need to divide the interval $\left[-\frac{\Tilde{K}}{2}, \frac{\Tilde{K}}{2}\right]$ into subintervals with length at most $\frac{\tilde{\epsilon}}{\sqrt{TD}}$, of which there are $\lceil \sqrt{TD} \Tilde{K} \tilde{\epsilon}^{-1} \rceil$ of them.
Further, because there are $TD$ such intervals to divide, the support of our code should be at least as large as the number of required subintervals as below:
\[ |\mathcal{V}|^M \geq \lceil \sqrt{TD} \Tilde{K} \tilde{\epsilon}^{-1} \rceil^{TD}  \] 

In the two cases below, we will attempt to construct this support by scaling the vocabulary size $|\mathcal{V}|$ and the code length $M$ individually. It is technically also possible to scale both at the same time, but we focus on the existence for this proof, and leave more efficient solutions to future work.

\textbf{Case 1} (Modality Tokenizer): Take $N = T$, and construct a vocabulary $\mathcal{V}$ with size $M_\epsilon = \lceil \sqrt{TD} \tilde{K} \tilde{\epsilon}^{-1} \rceil ^D$. We then define the encoding function for $x$ as $f_\epsilon(x) = (c_1, \dots, c_T)$ where $c_t \in \mathcal{V}$. We define the index of $c_t$ as a $D$-digit number of base-$M_\epsilon$
$$(d_1(c_t), d_2(c_t), \dots, d_D(c_t)),$$
where $d_n(c_t) = \lfloor \frac{x_{tn} + \tilde{K} / 2}{\tilde{\epsilon} / \sqrt{TD}} \rfloor$ is the $n$-th digit. Note that this is well defined because $d_n(c_t) \in \left[0, \lfloor \sqrt{TD} \tilde{K} \tilde{\epsilon}^{-1} \rfloor \right]$, and thus enables the representation of $\lceil \sqrt{TD} \tilde{K} \tilde{\epsilon}^{-1} \rceil^D$ unique values for $c_t$ using $D$ digits.
This way, given $f_\epsilon(x) = f_\epsilon(x')$, since $c_t = c_t'\ \forall t \in [1, T]$, we have $d_n(c_t) = d_n(c_t')\ \forall n \in [1, D]$, which implies that $|x_{tn} - x'_{tn}| \leq \tilde{\epsilon} / \sqrt{TD}$. Thus, we have
\begin{align}
\| x - x' \| = \sum_{t=1,d=1}^{T, D} (x_{td} - x'_{td})^2 \leq \sum_{t=1,d=1}^{T, D} \frac{\tilde{\epsilon}^2}{TD} = \epsilon
\end{align}
\textbf{Case 2} (Finding a Language Expression): Assume that the vocabulary has a fixed size $|\mathcal{V}| = N$ (which applies to a wide range of natural and machine languages). Take sequence length $N_\epsilon = TD \lceil \log_M \sqrt{TD} \tilde{K} \tilde{\epsilon}^{-1} \rceil$ and define 
$$f_\epsilon(\mathbf{x}) = (c_1, \dots, c_{N_\epsilon}),\ c_t \in \mathcal{V},$$
We map each $x_{td}$ to a number $u_{td} \in [0, 1]$ represented by $L$ digits of base-$N$, where the $n$-th digit $d_n(u_{td}) = c_{s + n}$. 
In this manner, $x_{td}$ will correspond to subsequence $(c_{s+1}, c_{s+2}, \dots, c_{s+L})$, where $s = (td - 1) \lceil \log_M \sqrt{TD} \tilde{K} \tilde{\epsilon}^{-1} \rceil$ and $L = \lceil \log_N \sqrt{TD} \tilde{K} \tilde{\epsilon}^{-1} \rceil$.
Thus, we have that
\begin{align}
x_{td} = \left( \sum_{n=1}^L \frac{d_n(u_{td})}{M^n} - \frac{1}{2} \right) \tilde{K}.
\end{align}
Now, given $\mathbf{x}, \mathbf{x}'$ where $f_\epsilon(\mathbf{x}) = f_\epsilon(\mathbf{x}')$, we have $f_\epsilon(\mathbf{x})_{td} = f_\epsilon(\mathbf{x}')_{td}$ and $d_n(u_{td}) = d_n(u_{td})$, so $u_{td}$ and $u'_{td}$ must lie in the same interval formed by numbers for which all digits $d_n$ are identical as below:
\begin{align} 
\left[ u:\ d_n(u) = v_n,\ n = 1, \dots, L \right] = \left[ 
\sum_{n=1}^L \frac{v_n}{M^n}, \sum_{n=1}^L \frac{v_n}{M^n} + \frac{1}{M^L} \right)
\end{align}
Thus we have
\begin{align*}
    |x_{td} - x_{td}'|
    &= \left| \left( \sum_{n=1}^L \frac{d_n(u_{td})}{M^n} - \frac{1}{2} \right) \tilde{K} - \left( \sum_{n=1}^L \frac{d_n(u_{td}')}{M^n} - \frac{1}{2} \right) \tilde{K} \right| 
    = \left| \left( \sum_{n=1}^L \frac{d_n(u_{td})}{M^n} - \sum_{n=1}^L \frac{d_n(u_{td}')}{M^n} \right) \tilde{K} \right| \\
    &\leq \left| M^{-L} \tilde{K} \right| = \left| M^{-\lceil \log_M \sqrt{TD} \tilde{K} \tilde{\epsilon}^{-1} \rceil} \tilde{K} \right| 
    \leq \left| M^{-\log_M \sqrt{TD} \tilde{K} \tilde{\epsilon}^{-1}} \tilde{K} \right| \leq \left| \frac{\tilde{\epsilon}}{\sqrt{TD}} \right|
\end{align*}
And by extension
$\| \mathbf{x} - \mathbf{x}' \| \leq \epsilon$
\end{proof}

\section{Proof for Proposition~\ref{prop:collapse-of-latent-loss}}
\label{appendix:proof-collapse-of-latent-loss}
\begin{proof}
    We will prove the proposition by constructing such a $(h^*, f^*)$ pair such that $\mathcal{L}_{\text{latent}}(h^*, f^*) = \min_{h, f} \mathcal{L}_{\text{latent}}(h, f)$. ($(h^*, f^*)$ is the global minimum of $\mathcal{L}_{\text{latent}}$.) \\
    This could be simply done by finding arbitrary $c \in \mathbb{R}^n$ and letting $h^*(o) = c$ for all $o \in \mathcal{O}$ and $f^*(c, a) = c$ for all $a \in \mathcal{A}$. Then, we have that $\mathcal{L}_{\text{latent}}(h^*, f^*) = 0$. Since $\mathcal{L}_{\text{latent}}$ is a distance measurement, we have that $\mathcal{L}_{\text{latent}} \geq 0$, which means that $(h^*, f^*)$ is the global minimum.
\end{proof}

\section{Proof for Proposition~\ref{prop:non-collapse-of-gen-loss}}
\label{appendix:proof-non-collapse-of-gen-loss}
\begin{proof}
We prove the proposition by showing that, with the generative loss $\mathcal{L}_{\text{gen}}$, there is no solution that can be both degenerate and optimal at the same time. More specifically, for any degenerate solution $(h', f')$, and fixed $g$, there must be a way to construct another solution $(\tilde{h}, \tilde{f})$ which yields a lower loss value. \\
More formally, without loss of generality, assume that we have:
\begin{itemize}
    \item Some arbitrary constant $c \in \mathbb{R}^n$;
    \item $h’: \mathcal{O} \rightarrow \mathcal{S}$ such that $h’(o) = c$ for all $o \in \mathcal{O}$;
    \item $f’: \mathcal{S} \times \mathcal{A} \rightarrow \mathcal{S}$ such that $f’(c, a) = c$ for all $a \in \mathcal{A}$;
    \item A fixed decoder $g: \mathcal{S} \rightarrow \mathcal{O}$;
    \item A dataset $\mathcal{D} \subset \mathcal{O} \times \mathcal{A} \times \mathcal{O}$ containing at least two distinct triplets:
    $$(o_1, a_1, o_2),\quad (o_3, a_3, o_4) \in \mathcal{D} \text{ with } o_2 \ne o_4.$$
\end{itemize}
Assuming all the function families are sufficiently expressive. Then, under the degenerate solution $(h’, f’)$, for any $(o, a, o’) \in \mathcal{D}$, we have:
$g \circ f'(h'(o), a) = g(f'(c, a)) = g(c).$
Therefore, the generative loss becomes:
\[
\mathcal{L}_{\text{gen}}(h', f', g) = \mathbb{E}_{(o, a, o') \sim \mathcal{D}} \left[ \left\lVert g(c) - o' \right\rVert \right].
\]
Note that this loss is non-zero if there exists at least one $(o, a, o’) \in \mathcal{D}$ such that $g(c) \ne o’$, which must be the case here because $o_2 \ne o_4$ while $g(c)$ is fixed across all inputs. \\
We now construct an alternative pair $(\tilde{h}, \tilde{f})$ such that:
\begin{itemize}
    \item For all $(o, a, o’) \in \mathcal{D} \setminus \{(o_1, a_1, o_2), (o_3, a_3, o_4)\}$, define $\tilde{h}(o) = c$ and $\tilde{f}(c, a) = c$, i.e., $\tilde{h}, \tilde{f}$ behave identically to the degenerate solution.
    \item For the two special cases, define $s_1, s_2 \in \mathcal{S}$ such that $g’(s_1) = o_2$ and $g’(s_2) = o_4$. (This is possible since $g’$ is fixed and assumed expressive enough to approximate $o_2$ and $o_4$.) \\
    Then, set $\tilde{h}(o_1) = \hat{s}_1$, $\tilde{f}(\hat{s}_1, a_1) = s_1$, and $\tilde{h}(o_3) = \hat{s}_2$, $\tilde{f}(\hat{s}_2, a_3) = s_2$, where $\hat{s}_1, \hat{s}_2 \in \mathcal{S}$ are any distinct intermediate representations to encode $o_1$ and $o_3$.
\end{itemize}
Now, consider the loss under this construction:
\begin{itemize}
    \item At $(o_1, a_1, o_2)$: $\tilde{h}(o_1) = \hat{s}_1$, $\tilde{f}(\hat{s}_1, a_1) = s_1$, $g’(s_1) = o_2$, so loss is $\left\lVert o_2 - o_2 \right\rVert = 0$. 
    \item At $(o_3, a_3, o_4)$: similarly, loss is zero for $\tilde{h}$ and $\tilde{f}$.
    \item While for $h'$ and $f'$, there must be one of $(o_1, a_1, o_2)$ and $(o_3, a_3, o_4)$ at which the loss is strictly positive. In other words, $\left\lVert g(c) - o_2 \right\rVert + \left\lVert g(c) - o_4 \right\rVert > 0$
    \item At all other datapoints, the loss is unchanged (equal to the degenerate solution’s loss) because we did not modify the outputs for those inputs.
\end{itemize}
Hence, the total loss under $(\tilde{h}, \tilde{f})$ is strictly less than under $(h’, f’)$:
\[\mathcal{L}_{\text{gen}}(\tilde{h}, \tilde{f}, g) < \mathcal{L}_{\text{gen}}(h', f', g)\]
This indicates that the degenerate solution must not be optimal, thus proving that it cannot be a global minimizer when $g$ is trained or expressive and when the dataset includes different outputs for different inputs.

\end{proof}

\section{Proof of Theorem~\ref{thm:latent-lower-bound-generative}}
\label{appendix:latent-lower-bound-generative}
\begin{proof}
Because the conditional distributions in the statement are all isotropic Gaussians, we can use them to construct the following transition distributions (assuming $\hat{q}$ is the empirical data distribution):
\begin{align*}
    \text{Latent Prediction } p_{h \circ f}:\ &\hat{s}' \mid o, a \sim \mathcal{N}(f(h(o), a), I) \\
    \text{Latent Target } \hat{q}_h:\ &\hat{s}' \mid o' \sim \mathcal{N}(h(o'), I) \\
    \text{Observation Prediction } p_{h \circ f \circ g}:\ &\hat{o}' \mid o, a \sim \mathcal{N}(g \circ f(h(o), a), I)
\end{align*}
It emerges that the two loss functions described above can be expressed as scaled KL divergences of those distributions, as shown below:
\begin{align*}
    \mathcal{L}_{\text{latent}} &= \| f(h(o), a) - h(o') \| = \sqrt{2 D_{\text{KL}}(\hat{q}_h(\hat{s}' \mid o') \, \Vert \, p_{h \circ f}(\hat{s}' \mid o, a))}, \\
    \mathcal{L}_{\text{gen}} &= \| g \circ f(h(o), a) - o' \| = \sqrt{2 D_{\text{KL}}(\hat{q}(o') \, \Vert \, p_{h \circ f \circ g}(o' \mid o, a))}.
\end{align*}
Consider applying the encoder transition to $\hat{q}$ and $p_{h \circ f \circ g}$. In the former case, we will recover the latent target $\hat{q}_h$, and in the latter case we will have performed a roundtrip from latent to observation, and back to latent, resulting in the following conditional distribution:
\begin{align*}
    \text{Roundtrip Latent Prediction } p_{h \circ f \circ g \circ h}:\ &\hat{s}' \mid o, a \sim \mathcal{N}(h \circ g \circ f(h(o), a), I).
\end{align*}
Applying the \textbf{data processing inequality} gives the following relation:
\[
    D_{\text{KL}}(\hat{q}_h(\hat{s}' \mid o') \, \Vert \, p_{h \circ f \circ g \circ h}(\hat{s}' \mid o, a)) \leq
    D_{\text{KL}}(\hat{q}(o') \, \Vert \, p_{h \circ f \circ g}(o' \mid o, a)),
\]
And because the left-hand side is $\frac{1}{2}\|h \circ g \circ f(h(o),a) - h(o')\|^2 $ and the right-hand side is $\frac{1}{2} \mathcal{L}_{\text{gen}}^2$, we plug them into the inequality above and get:
\begin{equation}
    \|h \circ g \circ f(h(o),a) - h(o')\| \leq \mathcal{L}_{\text{gen}}. 
    \label{ineq:data-processing-inequality}
\end{equation}
Next, because the encoder $h$ and decoder $g$ satisfy the roundtrip consistency of $\| h \circ g(\hat{s}) - \hat{s} \| \leq \epsilon$, plugging in $\hat{s} = f(h(o), a)$ gives:
\begin{equation}
    \| h \circ g \circ f(h(o), a) - f(h(o), a) \| \leq \epsilon.
    \label{ineq:roundtrip-consistency}
\end{equation}
Finally, we apply the triangle inequality to upper-bound the latent loss as below:
\begin{align*}
    \mathcal{L}_{\text{latent}} &= \| f(h(o), a) - h(o') \| \\
    &\leq {\underbrace {\textstyle \| f(h(o), a) - h \circ g \circ f(h(o), a) \|}_{\leq \epsilon \text{ by Inequality~\ref{ineq:roundtrip-consistency}}}} + 
        {\underbrace {\textstyle \| h \circ g \circ f(h(o), a) - h(o') \|}_{\leq \mathcal{L}_{\text{gen}} \text{ by Inequality~\ref{ineq:data-processing-inequality}}} } \\
    &\leq \mathcal{L}_{\text{gen}} + \epsilon
\end{align*}

When $h \circ g(\hat{s}) = \hat{s}$ for all $\hat{s} \in \mathcal{S}$, we have that $g: \mathcal{S} \to \text{Im}(g) \subseteq \mathcal{O}$ is an injective function which maps each point in $\mathcal{S}$ to a unique point in its image $\text{Im}(g)$. On the other hand, $h$ is the left inverse of $g$ on $\text{Im}(g)$, which maps each $\hat{o} \in \text{Im}(g)$ to $\hat{s} \in \mathcal{S}$ s.t. $g(\hat{s}) = \hat{o}$. Thus $h$ is bijective in $\text{Im}(g)$ by mapping each element $o \in \text{Im}(g)$ to exactly one $\hat{s} \in \mathcal{S}$.

This means $h$ defines a parameter transformation $T_h$ where $T_h(o) = h(o)$ for all $o \in \text{Im}(g)$. Because $\hat{q}(o')$ and $p_{h \circ f \circ g}(o' \mid o, a)$ are supported entirely inside $\text{Im}(g)$ and KL divergence is invariant under parameter transformation, we have:
\begin{align*}
    \frac{1}{2} \mathcal{L}_{\text{gen}}^2 &= D_{\text{KL}}(\hat{q}(o') \, \Vert \, p_{h \circ f \circ g}(o' \mid o, a)) \\
    &= D_{\text{KL}}(\hat{q}_h(\hat{s}' \mid o') \, \Vert \, p_{h \circ f \circ g \circ h}(\hat{s}' \mid o, a)) \\
    &= \frac{1}{2} \| h \circ g \circ f(h(o), a) - h(o') \|^2 \\
    &= \frac{1}{2} \| f(h(o), a) - h(o') \|^2 \quad (h \circ g(\hat{s}) = \hat{s}) \\
    &= \frac{1}{2} \mathcal{L}_{\text{latent}}^2.
\end{align*}
Thus we have $\mathcal{L}_{\text{latent}} = \mathcal{L}_{\text{gen}}$.
\end{proof}

\end{document}